\documentclass[a4paper,fleqn]{cas-dc}

\usepackage{listings}
\usepackage{amssymb}
\usepackage{hyperref}
\usepackage{graphicx}
\usepackage{multirow}
\usepackage{booktabs}
\usepackage{xurl}
\usepackage{lipsum}
\usepackage{lscape}
\usepackage{amsmath}
\usepackage{relsize}
\usepackage[figuresright]{rotating}
\usepackage{lineno}
\usepackage{algorithm}
\usepackage{algpseudocode}
\usepackage{comment}
\usepackage{nomencl}
\makenomenclature
\usepackage{calc}
\usepackage[T1]{fontenc}
\usepackage{subfigure}
\usepackage{tikz}
\usetikzlibrary{fit}
\usepackage{float}
\usepackage{orcidlink}
\usepackage{setspace}
\usepackage{tabularx}
\usepackage{array}
\usepackage{makecell}
\usepackage{natbib}
\usepackage{threeparttable}
\usepackage{booktabs}
\usepackage{tabularx}
\usepackage{array}
\usepackage{placeins}
\usepackage{float}
\usepackage{graphicx}
\usepackage{hyperref}
\setcitestyle{
  authoryear,
  open={(},
  close={)},
  citesep={;}
}

\hypersetup{
  colorlinks,
  citecolor=Violet,
  linkcolor=Red,
  urlcolor=Blue
}

\begin{document}
\raggedbottom    

\let\WriteBookmarks\relax


\shorttitle{Quantized Vision-Language Models for Plant Diagnosis}
\shortauthors{Wagle et al. (2026)}

\title[mode=title]{PlantVLM-Edge: Quantized Vision-Language Models for On-Device Crop Health Diagnosis}

\author[1]{Gaurab Wagle}\ead{oe.gaurab@gmail.com}
\author[2]{Bishal Adhikari}\ead{ba974@msstate.edu}
\author[3]{Jiajia Li}\ead{lijiajian@msu.edu}
\author[4]{Xiaolei Huang}\ead{xiaolei.huang@missouri.edu}
\author[2]{Dong Chen*}\ead{dchen@abe.msstate.edu}

\address[1]{Department of Computer Science and Engineering, Kathmandu University, 45210, Nepal}
\address[2]{Department of Agricultural and Biological Engineering, Mississippi State University, MS 39762, USA}
\address[3]{Electrical and Computer Engineering, Michigan State University, MI 48824, USA}
\address[4]{Electrical Engineering and Computer Science, University of Missouri, MO 65211, USA}
\address{*Dong Chen is the corresponding author}


\begin{abstract}
Timely and accurate crop health diagnosis is essential for sustainable agricultural production, yet accurate diagnosis often requires expert knowledge that is not readily available. Recent vision-language models (VLMs) have demonstrated remarkable capabilities in visual understanding, reasoning, and natural-language interaction, creating new opportunities for intelligent crop health assessment. However, deploying VLMs on resource-constrained devices remains challenging due to their substantial computational and memory requirements.
This paper presents PlantVLM-Edge, a quantized vision-language framework for conversational crop health diagnosis that can run entirely on resource-constrained edge devices. PlantVLM-Edge combines agricultural domain adaptation, post-training quantization, and on-device serving with an offline chat interface, allowing growers to ask open-ended questions and receive diagnoses, symptom explanations, and management guidance without cloud connectivity. The framework is backbone-agnostic: we validate it with two different VLM types with varying scale and architecture (Qwen3-VL-2B and LLaVA-1.5-7B), four quantization schemes, and two diagnostic domains (plant diseases and insect pests), and we deploy all quantized variants of both models on an NVIDIA Jetson AGX Orin. 
Eight-bit quantization leaves answer quality essentially unchanged relative to the full-precision model, and 4-bit activation-aware weight quantization (AWQ) retains 98.6\% of full-precision answer quality on both domains while reducing the weight memory of the 2B backbone by 49\%. In its most compact form, PlantVLM-Edge requires 2.01 GiB of weight memory, returns its first token in about 0.12 s, and completes a 64-token diagnostic response in 0.78 s. 
%
These results demonstrate that timely and accurate crop health diagnosis is practical on edge devices, providing a foundation for next-generation crop monitoring and decision support systems.
\end{abstract}



\maketitle


\section{Introduction}
Crop health underpins agricultural productivity and global food security, and protecting it depends on detecting threats early enough to act. Plant diseases and insect pest infestations are among the most persistent of these threats, accounting for annual crop yield losses of up to 40\% \citep{fao_plant_production, savary2019global}. Early and accurate diagnosis is essential to limiting economic damage, yet in practice it still depends heavily on trained agronomists or plant pathologists \citep{swanson2010strengthening}, a resource that is often scarce, expensive, or simply inaccessible in smallholder and remote farming contexts. This gap has motivated research into automated and scalable approaches to crop health monitoring that can bring expert-level diagnosis to growers without direct access to specialists.

Deep learning-based computer vision has made image-based recognition of plant diseases and pests highly accurate on benchmark datasets~\citep{singh2020review, liu2021plant}, first with convolutional neural network classifiers~\citep{mohanty2016using, ferentinos2018deep} and more recently with transformer-based classifiers and detectors~\citep{meng2023known, tan2025m2detr}. However, models trained on curated images often generalize less well to field
conditions~\citep{singh2020plantdoc}, and more fundamentally, they formulate diagnosis as a fixed recognition task. Their output is a class label or a bounding box, with no explanation of the visual evidence behind it and no way to ask follow-up questions. For non-expert growers, an isolated label is a major usability barrier, because it does not explain what the problem means or what to do about it. Addressing this requires models that go beyond passive detection to provide flexible, interactive decision support~\citep{novaliendry2026edge}.

Vision-language models (VLMs), which couple a pretrained visual encoder with a language model~\citep{ghosh2024exploring, zhao2026survey}, offer a path toward such support. A single VLM can answer free-form questions about an image, from identifying a crop or pest to explaining likely causes and recommending management actions, and it can be adapted to a new domain by fine-tuning on question-answer data~\citep{liu2023visual, li2024foundation}. Agricultural studies have adapted VLMs to crop disease and pest diagnosis through instruction tuning and parameter-efficient fine-tuning~\citep{awais2025agrogpt, wang2024agri, sharma2024llava, kaur2025visual, jena2026bananavlm}, and benchmarks such as CDDM~\citep{liu2024multimodal}, AgMMU~\citep{gauba2026agmmu}, IP-VQA~\citep{jin2025ip}, and PlantExpertVQA~\citep{sakib2026visual} now evaluate diagnostic question answering beyond fixed labels. These studies consistently show that general-purpose VLMs improve substantially once adapted to the agricultural domain (Section~\ref{sec:related-vfm}).

However, field deployments may prevent directly adopting large VLM toolkits due to resource constraints. In rural farming environments, unreliable internet connectivity, recurring cloud service costs, and data privacy concerns make on-device inference highly desirable \citep{yao2025efficient, yu2025cloud}. VLMs typically contain billions of parameters, which makes local execution expensive, and memory and speed remain limiting on edge devices. When run with llama.cpp inference framework, with its language model quantized to 4 bits, the general-purpose LLaVA-1.5-7B ~\citep{liu2024improved} still occupied 4.7 GB and took about 20 s to answer a single question on an NVIDIA Jetson AGX Orin and about 330 s on a smartphone ~\citep{chu2023mobilevlm}, delays that are impractical for a grower awaiting a diagnosis. 

%
Post-training quantization, which lowers the numerical precision of a model without retraining it, is a natural way to reduce these costs. However, existing edge evaluations of VLM quantization have used general-purpose models and benchmarks, and report that its speed benefits depend on the hardware and inference software (Section~\ref{sec:related-edge}). How quantization affects the answer quality, memory footprint, and latency of domain-adapted agricultural VLMs on embedded hardware therefore remains largely unexamined, as does whether a compact VLM can match a larger one once both are adapted and compressed.

To address this gap, we present PlantVLM-Edge, a quantized vision-language framework for conversational crop health diagnosis that runs entirely on edge devices. PlantVLM-Edge consists of three stages: agricultural domain adaptation of a pretrained VLM through parameter-efficient fine-tuning, post-training quantization of the adapted model, and on-device serving with an offline chat interface that growers can access from a web browser on a local network. The main contributions of this work are as follows:
\begin{itemize}
    \item We develop PlantVLM-Edge, an end-to-end framework that takes a general-purpose VLM from domain adaptation to fully offline deployment, allowing growers to ask open-ended questions and receive diagnoses, symptom explanations, and management guidance without cloud connectivity.
    \item We show that the framework is backbone-agnostic by instantiating it with two VLMs of different scale and architecture, Qwen3-VL-2B and LLaVA-1.5-7B, across two diagnostic domains: plant diseases and insect pests.
    \item We systematically evaluate four post-training quantization configurations, spanning round-to-nearest, AWQ, and SmoothQuant at 8-bit and 4-bit precision. We measure both answer quality and on-device efficiency (weight memory, time to first token, inter-token latency, and end-to-end latency) on an NVIDIA Jetson AGX Orin.
    \item We provide practical deployment guidance. Eight-bit quantization leaves answer quality essentially unchanged, 4-bit AWQ retains 98.6\% of full-precision answer quality while reducing weight memory by 49\%, and the choice of backbone affects deployment cost more than the choice of quantization method.
\end{itemize}
The remainder of this paper is organized as follows. Section~\ref{sec:related-work} reviews related work, Section~\ref{sec:methodology} describes the datasets, the PlantVLM-Edge framework, and the evaluation protocol, Section~\ref{sec:results} presents the results, Section~\ref{sec:discussion} discusses their implications and limitations, and Section~\ref{sec:conclusion} concludes the paper.

\section{Related Works}
\label{sec:related-work}
\subsection{Computer vision for crop health diagnosis}
Deep learning, and computer vision in particular, has become the dominant approach to automated crop disease and pest recognition. Early methods relied on handcrafted visual descriptors such as color, texture, and shape, combined with classical machine-learning classifiers such as support vector machines and random forests \citep{sladojevic2016deep}. Subsequently, convolutional neural networks (CNNs), which learn hierarchical visual features directly from images, substantially outperformed these approaches once large labeled datasets became available. For example, Mohanty et al. \citep{mohanty2016using} trained CNNs on 54,306 PlantVillage images covering 14 crop species and 26 diseases, achieving 99.35\% accuracy, and Ferentinos \citep{ferentinos2018deep} extended this to 58 plant-disease classes across 25 crops with a best accuracy of 99.53\%. However, these results were obtained on images captured under controlled conditions. When Mohanty et al. tested their model on images from other sources, accuracy fell to about 31\%. Consequently, later datasets such as PlantDoc \citep{singh2020plantdoc} were introduced specifically to capture real-world field conditions. To address this challenge, subsequent work improved robustness through transfer learning, data augmentation, and field-acquired training images \citep{thenmozhi2019crop, xu2022transfer, gui2021towards}. CNN-based object detectors, particularly the YOLO family, have also been widely adopted for localizing diseased regions and pests in real time, including on embedded hardware \citep{zhang2024gvc}.

More recently, transformer-based architectures have been adopted for crop health diagnosis. The vision transformer (ViT) \citep{dosovitskiy2020image} divides an image into patches and uses self-attention to model relationships across the entire image, capturing global context that the local receptive fields of CNNs can miss. Pretraining on large plant image collections has proven particularly effective. For example, a ViT pretrained on PlantCLEF2022 achieved strong closed-set accuracy across four plant disease datasets and could also recognize disease classes unseen during training \citep{meng2023known}. For detection, transformer-based detectors following DETR \citep{carion2020end} remove hand-designed components such as anchor boxes and non-maximum suppression. M2DETR \citep{tan2025m2detr} adapted this design to pest detection with a multi-scale denoising encoder, outperforming both CNN-based and transformer-based detectors on the IP102 pest dataset \citep{wu2019ip102}. Despite these architectural advances, both CNN and transformer-based approaches formulate crop health diagnosis as a fixed recognition or detection problem: their output is a class label, a bounding box, or a segmented region, with no explanation of the underlying visual evidence and no support for follow-up questions. This limitation motivates the use of vision-language models, reviewed in the next subsection.

\subsection{Vision foundation models in agriculture}
\label{sec:related-vfm}
Foundation models are large neural networks trained on broad data at scale that can be adapted to a wide range of downstream tasks \citep{li2024foundation}. Their strength comes from the scale of pretraining. By learning from vast and diverse data, typically with self-supervised objectives that require no manual labels, they acquire general-purpose representations and knowledge that transfer across tasks and domains. As a result, a single pretrained model can be adapted to new tasks with far less labeled data than training from scratch would require, and in many cases it can perform tasks it was never explicitly trained for. Large language models (LLMs) illustrate this most clearly. GPT-3, with 175 billion parameters, performed a variety of language tasks from only a few examples given in its prompt, without task-specific fine-tuning \citep{brown2020language}. Parameter-efficient methods such as LoRA \citep{hu2021lora} have further lowered the cost of adaptation by updating only a small fraction of a model's weights.

The same paradigm has been extended to computer vision. CLIP \citep{radford2021learning} learned to align images and text by training on 400 million image-caption pairs collected from the web, enabling it to classify images into categories described in natural language without task-specific training. The Segment Anything Model (SAM) \citep{kirillov2023segment} was trained on more than one billion masks across 11 million images and can segment objects in images it has never seen, and DINOv2 \citep{oquab2023dinov2} showed that self-supervised pretraining alone can produce visual features that perform well across many downstream tasks. Vision-language models (VLMs) combine these advances with LLMs by connecting a pretrained visual encoder to a language model through a lightweight projector. Models such as BLIP-2 \citep{li2023blip}, LLaVA \citep{liu2023visual}, and Qwen-VL \citep{qwen3vltechnicalreport} can answer free-form questions about images, describe visual content, and reason about what they see in natural language, which makes them the most direct foundation-model counterpart to an interactive diagnostic assistant.

These capabilities have attracted growing interest in agriculture, where labeled data is costly to collect and diagnostic questions go beyond fixed categories. Vision foundation models have been applied to agricultural image analysis tasks such as segmentation and phenotyping \citep{li2024foundation}, while most work on crop health diagnosis has focused on adapting VLMs. AgroGPT \citep{awais2025agrogpt} used domain-specific instruction tuning to enable a general-purpose multimodal model to answer fine-grained questions about crops and diseases, and Agri-LLaVA \citep{wang2024agri} introduced knowledge-infused training to build a multimodal assistant covering agricultural pests and diseases. Other studies have specialized VLMs for individual crops, including a BLIP-2-based model for tomato leaf disease diagnosis \citep{kaur2025visual} and BananaVLM \citep{jena2026bananavlm}, built on LLaVA-1.5-7B, for banana diseases, while LLaVA-PlantDiag \citep{sharma2024llava} explored conversational plant pathology diagnosis. A common finding is that general-purpose VLMs perform poorly on agricultural tasks without adaptation but improve substantially after domain fine-tuning. On the CDDM benchmark, for instance, fine-tuning raised the disease classification accuracy of LLaVA-1.5-7B from 5.9\% to 91.8\% \citep{liu2024multimodal}. Benchmarks such as CDDM, AgMMU \citep{gauba2026agmmu}, IP-VQA \citep{jin2025ip}, and PlantExpertVQA \citep{sakib2026visual} now support evaluation beyond fixed labels, covering species identification, symptom description, causal reasoning, and pest management. However, this body of work has focused primarily on adaptation and diagnostic quality, typically assuming access to server-class GPUs. How these models can be deployed efficiently on edge hardware, where agricultural diagnosis is ultimately needed, has received far less attention, as reviewed in the next subsection.

\subsection{Edge AI for agricultural applications}
\label{sec:related-edge}
Edge AI refers to running inference directly on devices located near where data is collected, rather than sending data to remote servers \citep{el2024leveraging}. In agriculture, this approach addresses several practical constraints: internet connectivity in rural areas is often limited or unreliable, field decisions benefit from low-latency responses, and farm data can remain under the grower’s control. Embedded platforms such as the NVIDIA Jetson family and the Raspberry Pi, along with smartphones, have therefore become common targets for agricultural vision systems. Most deployments to date have used CNN-based models, often made practical through lightweight architectures such as MobileNets \citep{howard2017mobilenets} and model compression \citep{cheng2018model}. For example, YOLO-based detection of aphid-damaged cotton leaves has been run in real time on an NVIDIA Jetson Xavier NX \citep{zhang2024gvc}, and a portable plant disease detection system has been deployed on a Jetson Nano \citep{yousif2025portable}. \cite{adhikari2025comprehensive} evaluated YOLO-based deer detection across multiple edge devices, including the Raspberry Pi and Jetson AGX Xavier, and UAV-based systems have integrated embedded hardware for on-site pest detection to guide precision spraying \citep{chen2021identification}. These studies show that edge deployment of conventional vision models is mature, but the deployed models produce fixed outputs such as class labels and bounding boxes.

Deploying VLMs on edge hardware is considerably more challenging. A VLM must encode the image with a vision encoder and then generate its answer token by token with a language model containing billions of parameters, so its memory and latency requirements far exceed those of a CNN classifier or detector. Recent work has begun to address this. MobileVLM \citep{chu2023mobilevlm} designed compact VLMs for mobile devices and benchmarked them alongside LLaVA-1.5 on a smartphone and an NVIDIA Jetson AGX Orin, with the language models quantized to 4-bit. Shin et al. \citep{shin2026rethinking} evaluated component-wise quantization of several small VLMs, including Qwen3-VL-2B, on Jetson Orin NX and AGX devices, and found that 4-bit quantization greatly reduced memory use but did not necessarily speed up generation, with efficiency depending strongly on the hardware platform. On the software side, post-training quantization methods such as GPTQ \citep{frantar2022gptq}, AWQ \citep{lin2024awq}, and SmoothQuant \citep{xiao2023smoothquant}, together with serving engines such as vLLM \citep{kwon2023efficient}, have made it practical to compress and serve these models without retraining. In agriculture, Kaur et al. \citep{kaur2025visual} developed a BLIP-2-based tomato disease model targeting a portable greenhouse monitoring device, one of the few agricultural VLM studies to consider deployment hardware.

These studies leave an important gap. Edge evaluations of VLMs have focused on general-purpose models and benchmarks rather than domain-adapted agricultural models, while agricultural VLM studies have focused on diagnostic quality rather than deployment efficiency. As a result, the combined effect of post-training quantization on the answer quality, memory footprint, and latency of agricultural VLMs running on embedded hardware remains insufficiently characterized, and it is unclear which models and quantization configurations offer the best trade-off for field use. This work addresses that gap by building and evaluating a complete pipeline, from domain adaptation through quantization to offline conversational deployment, on an NVIDIA Jetson AGX Orin.

\section{Methods}
\label{sec:methodology}
This section first describes the datasets used for training and evaluation (Section~\ref{sec:datasets}). It then presents the PlantVLM-Edge framework and each of its components (Section~\ref{sec:PlantVLM-Edge}), and finally defines the protocol and metrics used to evaluate answer quality and on-device efficiency (Section~\ref{sec:evaluation}).

\begin{figure*}[H]
\centering
\includegraphics[width=0.9\linewidth]{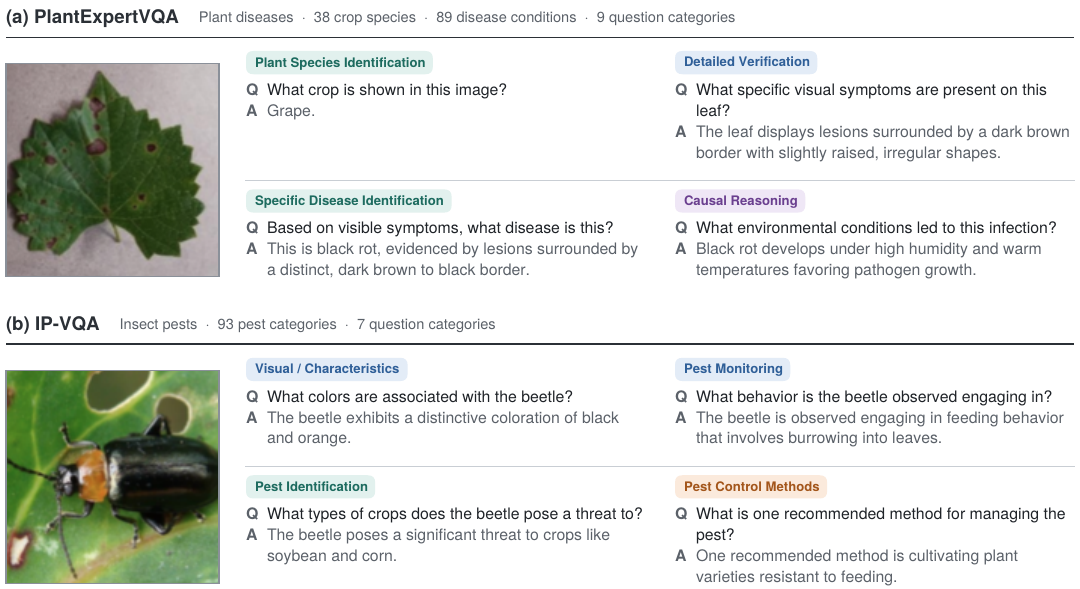}
\caption{Representative image-question-answer pairs from the PlantExpertVQA and IP-VQA datasets.}
\label{fig:dataset_vqa_examples}
\end{figure*}

\subsection{Datasets}
\label{sec:datasets}
We derive our datasets from two publicly available agricultural visual question answering (VQA) datasets: PlantExpertVQA \citep{sakib2026visual} for plant disease diagnosis and IP-VQA \citep{jin2025ip} for insect pest identification. Because PlantExpertVQA is too large to train on in full with our computational resources, we draw a stratified subset of it. IP-VQA is small enough to use in full after cleaning, and we group its questions into seven categories for evaluation. Figure~\ref{fig:dataset_vqa_examples} shows representative image-question-answer examples from both datasets, illustrating the range of questions they cover, from crop and pest identification to symptom description, causal reasoning, and pest management.

\paragraph{PlantExpertVQA} \citep{sakib2026visual} is a large-scale plant disease VQA dataset comprising 765,186 expert-verified question-answer pairs over 150,841 images. It spans 38 crop species and 89 disease conditions, and its questions are organized into nine categories across three levels of cognitive complexity. The dataset is released with an image-disjoint 70/10/20 split into training, validation, and test partitions.

Training on the full corpus would have required substantially more computational time and resources, so we constructed a stratified training subset from the authors’ released training partition. Sampling was performed at the image level, so that all question-answer pairs belonging to a selected image were retained together and the splits remained image-disjoint. Strata were defined by the combination of crop species, disease condition, and source dataset, and images were allocated to strata in proportion to their size using the largest-remainder method. We also cropped the static black borders present in some images so that the model trained only on relevant visual content. The resulting training subset contains 63,546 question-answer pairs over 12,564 images, approximately 8\% of the full dataset.

For evaluation, we used a stratified subset of 3,000 images from the released test partition, comprising 15,082 question-answer pairs. A separate set of 1,500 question-answer pairs from the validation partition was used for validation during training. Table~\ref{tab:plantexpert_categories} lists the number of evaluation pairs in each question category.

\begin{table}[!t]
\centering
\small
\renewcommand{\arraystretch}{1.15}
\setlength{\tabcolsep}{10pt}
\caption{Question categories in the PlantExpertVQA evaluation set, where $n$ is the number of question-answer pairs in each category.}
\label{tab:plantexpert_categories}
\begin{tabular}{@{}l r@{}}
\toprule
\textbf{Question Category} & \textbf{n} \\
\midrule
Existence \& sanity check       & 72 \\
Plant species identification    & 2{,}147 \\
General health assessment       & 1{,}337 \\
Visual attribute grounding      & 716 \\
Detailed verification           & 1{,}689 \\
Specific disease identification & 3{,}972 \\
Comprehensive description       & 1{,}767 \\
Causal reasoning                & 2{,}180 \\
Counterfactual reasoning        & 1{,}202 \\
\bottomrule
\end{tabular}
\end{table}

\paragraph{IP-VQA} \citep{jin2025ip} is an insect pest VQA dataset covering 102 pest species, with approximately five question-answer pairs per image. Its questions address pest characteristics, pest categories, crop damage, and pest management practices. We removed pest classes with fewer than 10 images, to exclude severely underrepresented classes, and then removed corrupted images. This left 18,950 images and 94,785 question-answer pairs across 93 pest categories, which we divided into image-disjoint training (70\%), validation (10\%), and test (20\%) sets. For category-level analysis, we grouped questions into seven evaluation categories by keyword matching on the question text; questions that matched no pattern were grouped as ``Others''. Table~\ref{tab:ipvqa_categories} summarizes these categories.
   
\begin{table}[!t]
\centering
    \small
    \caption{Question categories in the IP-VQA evaluation set. Values in parentheses are the number of question-answer pairs in each category.}
    \label{tab:ipvqa_categories}
    \begin{tabularx}{\linewidth}{l X}
    \toprule
    \textbf{Category } & \textbf{Question focus} \\
        \midrule
            Pest identification & Pest identity and type (4354) \\
            Pest feeding and damage & Feeding, damage, and impact (4135)\\
            Optimal timing  & Timing and season of control (345) \\
            Pest monitoring & Monitoring and detection (1121)\\
            Pest control methods & Control, treatment, and prevention (3234) \\
            Visual / characteristics & Appearance and visual characteristics (2199) \\
            Others & Unmatched question types (3562) \\
        \bottomrule
\end{tabularx}
\end{table}

\begin{figure*}[t]
\centering
\includegraphics[width=\textwidth]{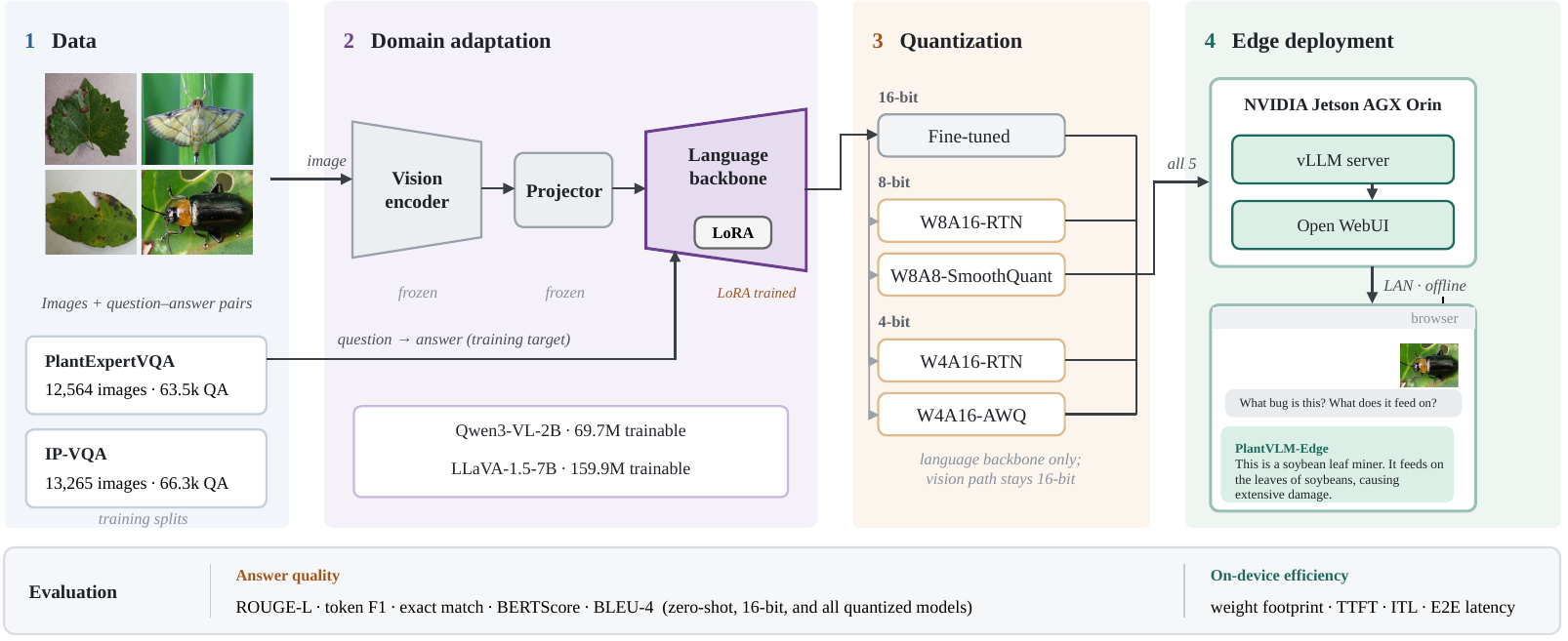}
\caption{Overview of the PlantVLM-Edge framework. A pretrained vision-language model is adapted to plant-disease and insect-pest question answering with LoRA, with the vision encoder and projector kept frozen. The fine-tuned model is then quantized into four configurations and served with vLLM on an NVIDIA Jetson AGX Orin, where its weight footprint and latency are measured. Growers interact with the deployed model offline through a browser-based chat interface.}
\label{fig:pipeline}
\end{figure*}


\subsection{PlantVLM-Edge framework}
\label{sec:PlantVLM-Edge}
Figure~\ref{fig:pipeline} provides an overview of PlantVLM-Edge. The framework takes a general-purpose pretrained VLM and prepares it for offline crop health diagnosis in three stages. In the domain adaptation stage, we fine-tuned the VLMs with LoRA on agricultural question-answer data. 
In the compression stage, the merged model is quantized after training into lower-precision formats that reduce its memory footprint and accelerate text generation. In the deployment stage, the quantized model is served with vLLM on an NVIDIA Jetson AGX Orin and accessed through a browser-based chat interface, allowing users on the same local network to query it without an internet connection. The framework is backbone-agnostic: any VLM supported by the fine-tuning, quantization, and serving toolchains can be used.

\subsubsection{Vision-language-model backbones}
\label{sec:models}
We instantiate PlantVLM-Edge with two open-weight, instruction-tuned VLMs
available on the Hugging Face Hub: Qwen3-VL-2B Instruct~\citep{qwen3vltechnicalreport} and
LLaVA-1.5-7B~\citep{liu2024improved}. Both follow the standard VLM design of a vision encoder, a lightweight projector, and a language backbone, but they differ in architecture and pretraining. In particular, Qwen3-VL processes images at dynamic resolution, whereas LLaVA-1.5 uses a fixed-size CLIP input, so the two produce different numbers of visual tokens per image.
 
The two models were selected for three reasons. First, both are open-weight and reproducible, yet differ in architecture. Second, both are supported end-to-end by the \texttt{llmcompressor} quantization toolkit \citep{llmcompressor2024} and the vLLM serving engine \citep{kwon2023efficient}, which makes the full pipeline feasible. Third, they span the range of model sizes that can be deployed on our target device: Qwen3-VL-2B has approximately 2~billion parameters (a 16-bit weight footprint of 3.96~GiB), whereas LLaVA-1.5-7B has approximately 7~billion parameters (13.16~GiB). The two models therefore represent realistic deployment candidates at different points in the size range that fits on the target device, rather than a controlled comparison of model size: because they also differ in architecture and pretraining, our comparison addresses whether a compact model is sufficient for deployment, not how model size alone affects performance.

\subsubsection{Domain adaptation}
\label{sec:finetuning}
The first stage of PlantVLM-Edge adapts the pretrained backbone to agricultural diagnosis. To efficiently specialize the models while preserving their pretrained knowledge, both models were adapted to each domain using Low-Rank Adaptation (LoRA)~\citep{hu2021lora} applied to the language backbone, while the vision encoder, multimodal projector, token embeddings, and output head remained frozen. Specifically, adapters were inserted into seven projection layers: query, key, value, and output in the attention modules, and gate, up, and down in the MLP modules.

The same optimization configuration (Table~\ref{tab:finetune-config}) was used for all four fine-tuning runs (two models $\times$ two datasets), and each model used its own native chat template. Furthermore, we trained for at most three epochs with early stopping on validation loss (patience of three evaluations) and kept the checkpoint with the lowest validation loss. All runs used a single NVIDIA A100-SXM4-80GB GPU. After training, the LoRA adapters were merged into the language backbone to produce a single fine-tuned checkpoint for quantization~\citep{hu2021lora}.
 
For image processing, images were kept in their original stored form and processed online by each model's image processor. For Qwen3-VL, dynamic-resolution processing was configured to produce between 128 and 256 visual tokens per image. In contrast, LLaVA-1.5 used its standard CLIP image processor, which resizes the shortest image edge to 336~pixels and applies a $336 \times 336$ center crop during both training and inference, producing 576 visual tokens.
    
\begin{table}[H]
\centering
\caption{Fine-tuning configuration. Optimization settings are shared across both models and all fine-tuning runs; model-specific settings are listed separately.}
\label{tab:finetune-config}
\footnotesize
\setlength{\tabcolsep}{5pt}
\begin{tabular}{@{}lcc@{}}
\toprule
 & \textbf{Qwen3-VL-2B} & \textbf{LLaVA-1.5-7B} \\
\midrule
\multicolumn{3}{@{}l}{\textit{LoRA}}\\
Rank $r$              & 64      & 64 \\
$\alpha$             & 128     & 128 \\
Dropout              & 0.05    & 0.05 \\
Trainable params     & 69.73M   & 159.91M \\
\% of model          & 3.17\%  & 2.21\% \\  
\midrule
\multicolumn{3}{@{}l}{\textit{Precision \& tokens}}\\
Compute precision    & BF16    & FP16 \\
Max sequence length  & 1{,}024 & 1{,}024 \\  
Visual tokens/image  & 128-256 & 576 \\
\midrule
\multicolumn{3}{@{}l}{\textit{Optimisation (shared)}}\\
Optimiser        & \multicolumn{2}{c}{AdamW ($\beta_1{=}0.9,\ \beta_2{=}0.999,\ \epsilon{=}10^{-8}$)} \\
Learning rate    & \multicolumn{2}{c}{$1\times10^{-4}$ (cosine)} \\
Effective batch  & \multicolumn{2}{c}{16 ($4\times4$ accum.)} \\
Grad.\ checkpointing & \multicolumn{2}{c}{Enabled} \\
Max epochs       & \multicolumn{2}{c}{3 (early stop, patience 3)} \\
GPU             & \multicolumn{2}{c}{One NVIDIA A100 80GB} \\
\bottomrule
\end{tabular}
\end{table}

\subsubsection{Post-training quantization}
\label{sec:quantization}
The second stage compresses the adapted model for edge deployment. After the LoRA adapters were merged, each fine-tuned model was quantized with \texttt{llmcompressor}~\citep{llmcompressor2024} into four configurations spanning three techniques. Each configuration is named by the precision of its weights and activations: W8A16, for example, denotes 8-bit weights and 16-bit activations, and the suffix gives the quantization method. The configurations therefore differ along three dimensions: weight precision (8 or 4 bits), whether activations are also quantized, and whether calibration data are required. In every configuration, quantization was restricted to the linear layers of the language backbone, which hold most of the model's parameters and must be read from memory at every decoding step~\citep{lin2024awq}. The vision encoder, multimodal projector, token embeddings, output head, and normalization weights were kept at 16-bit precision (BF16 for Qwen3-VL and FP16 for LLaVA), so that the visual features passed to the language backbone were identical across
configurations.

\paragraph{Round-to-Nearest (RTN)}
RTN is the simplest form of post-training quantization~\citep{nagel2021white}. For each group of weights, it computes a scale from the largest absolute weight in the group, divides every weight by this scale, and rounds the result to the nearest integer that can be represented at the target bit-width. Because the scales are derived from the weights alone, RTN requires no calibration data: it never runs example inputs through the model to decide how to quantize. We apply RTN at two bit-widths. \emph{W8A16-RTN} uses 8-bit weights with one scale per output channel, and \emph{W4A16-RTN} uses 4-bit weights with one scale per group of 128 weights; smaller groups give finer scales and therefore smaller rounding errors, which matters more at 4 bits. As the fastest and simplest method, RTN serves as the baseline against which the calibrated methods are compared.

\paragraph{Activation-Aware Weight Quantization (AWQ)}
At 4 bits, rounding errors become large enough to affect the model's outputs. AWQ~\citep{lin2024awq} reduces this effect by recognizing that a small fraction of weight channels matters far more than the rest, namely those that multiply inputs with large activation magnitudes. It runs a small calibration set through the model to identify these salient channels and scales them up before rounding, so that they are quantized with less relative error. Like RTN, AWQ changes only how weights are stored, so the quantized model runs with the same 4-bit inference kernels and memory footprint. Our \emph{W4A16-AWQ} configuration uses 4-bit weights with a group size of 128, calibrated on 128 question-answer examples drawn from 128 distinct training images.

\paragraph{SmoothQuant}
The weight-only methods above reduce memory but still perform all computation in 16 bits. SmoothQuant~\citep{xiao2023smoothquant} also quantizes the activations, allowing matrix multiplications to run with 8-bit integers. Activations are harder to quantize than weights because a few channels contain large outliers, which would force a coarse scale onto all other values. SmoothQuant therefore first divides each activation channel by a smoothing factor and multiplies the corresponding weights by the same factor, which leaves the layer output unchanged but shifts part of the quantization difficulty from the activations to the weights. With a smoothing strength of 0.5, the difficulty is split evenly between the two. Both are then quantized to 8 bits, using dynamic per-token scales for the activations. \emph{W8A8-SmoothQuant} was calibrated with the same 128-example protocol as AWQ.

Table~\ref{tab:quant-configs} summarizes the four configurations and the resulting weight footprint of each model. Quantization reduced the footprint substantially, and configurations at the same weight precision have identical sizes regardless of method, because the calibrated methods change which values are stored rather than how many bits are used. The relative reduction is larger at 4 bits and for the larger model, because the components kept at 16-bit precision make up a larger share of the compact model.

\begin{table}[!t]
\centering
\caption{Quantization configurations and the resulting weight footprint (GiB; 1~GiB $=$ 1.07~GB). All configurations quantize only the linear layers of the language backbone; the vision pathway remains at 16-bit precision.}
\label{tab:quant-configs}
\footnotesize
\setlength{\tabcolsep}{3pt}
\begin{threeparttable}
\begin{tabular*}{\columnwidth}{@{\extracolsep{\fill}}l l l l r r@{}}
\toprule
 & & & & \multicolumn{2}{c}{\textbf{Size (GiB)}} \\
\cmidrule(l){5-6}
\textbf{Config.} & \textbf{Weights} & \textbf{Act.} & \textbf{Calib.}
 & \textbf{Qwen} & \textbf{LLaVA} \\
\midrule
Fine-tuned & 16            & 16 & -      & 3.96 & 13.16 \\
W8A16-RTN  & 8 (per-ch.)   & 16 & None    & 2.65 & 7.13 \\
W8A8-SQ    & 8 (per-ch.)   & 8  & 128 ex. & 2.65 & 7.13 \\
W4A16-RTN  & 4 (g128)      & 16 & None    & 2.01 & 4.20 \\
W4A16-AWQ  & 4 (g128)      & 16 & 128 ex. & 2.01 & 4.20 \\
\bottomrule
\end{tabular*}
\begin{tablenotes}[flushleft]\scriptsize
  \item Weights and Act.: bit-width of weights and activations (8-bit activations use dynamic per-token scales); SQ: SmoothQuant; Calib.: calibration data; per-ch.: one scale per output channel; g128: one scale per group of 128 weights; 128 ex.: 128 calibration examples. Qwen: Qwen3-VL-2B (BF16); LLaVA: LLaVA-1.5-7B (FP16).
\end{tablenotes}
\end{threeparttable}
\end{table}

\subsubsection{Edge deployment and serving}
\label{sec:deployment}
The final stage serves the quantized model on the edge device and makes it accessible to users. Our target device is an NVIDIA Jetson AGX Orin Developer Kit, with a 12-core ARM Cortex-A78AE CPU at 2.2~GHz and 64~GB of memory shared between the CPU and GPU. A device of this class is the intended field platform: it requires no network connection to answer queries and keeps all data on the device. Because the memory is shared, it must hold the model weights, the key-value (KV) cache, and the serving runtime at the same time; reducing the space taken by the weights is precisely what quantization is intended to achieve.
 
To support real-time inference, each quantized checkpoint is served using vLLM~\citep{kwon2023efficient}, which provides efficient KV-cache management and native support for the quantized formats generated by \texttt{llmcompressor}. Specifically, vLLM runs inside a container built for the Jetson's CUDA~12.6 Tegra runtime. The model checkpoint is copied to the device, mounted within the container, and exposed through an OpenAI-compatible HTTP endpoint. 

To provide a user-friendly interface, Open WebUI~\citep{baek2025open} runs on the same device and communicates directly with the local vLLM server. Rather than interacting with the HTTP endpoint itself, users access the system through a browser-based chat interface that supports conversation history and voice input. As a result, anyone connected to the same local network can use the system through a standard web browser without requiring internet access. This fully offline architecture makes PlantVLM-Edge practical for field deployment and accessible to growers with little or no technical expertise.

\subsection{Evaluation}
\label{sec:evaluation}
We evaluate PlantVLM-Edge along two dimensions: the quality of its answers, measured against the reference answers of each dataset (Section~\ref{sec:answer-quality}), and its efficiency on the edge device, measured on the Jetson AGX Orin (Section~\ref{sec:efficiency}).

\subsubsection{Answer quality}
\label{sec:answer-quality}
Because the models generate free-form answers rather than selecting from fixed labels, no single metric fully captures answer quality. We therefore compare each generated answer with the dataset's single reference answer using five complementary metrics, ordered from the strictest to the most lenient: normalized exact match (EM), which requires the complete answer to match; token-level F1, which measures word overlap regardless of order; ROUGE-L F1, which rewards shared words that appear in the same order; corpus BLEU-4, which measures the overlap of short word sequences; and BERTScore F1, which measures semantic similarity even when the wording differs. These metrics are widely used to evaluate generated answers in agricultural VQA and related tasks~\citep{sakib2026visual, chen2019evaluating, wang2020asking}. We use ROUGE-L as the primary metric throughout the paper, because it gives credit for correct answers phrased differently from the reference while remaining sensitive to word order, and because it is also reported by the authors of PlantExpertVQA \citep{sakib2026visual}, which allows a comparison with their results.

The same scoring procedure was applied to both models on both datasets. Only the generated answer was scored: prompt tokens and special tokens were excluded, leading and trailing whitespace was removed, and neither the question nor the image was used by the scoring functions. For EM, token F1, and ROUGE-L, predictions and references were first normalized by lowercasing, removing ASCII punctuation, removing the articles \textit{a}, \textit{an}, and \textit{the}, and collapsing whitespace. BLEU and BERTScore were computed on the unnormalized text with their own tokenizers. All metrics are reported on a 0-100 scale. BLEU is computed once over the whole evaluation set, whereas the other four metrics are computed for each question and averaged, so that every question contributes equally.

\paragraph{Normalized exact match (EM)}
EM is the strictest metric: an answer scores only if it is identical to the reference after normalization \citep{rajpurkar2016squad}. Let $h_i$ and $r_i$ denote the predicted and reference answers for question $i$, and let $\mathcal{N}(\cdot)$ denote the normalization function. For $N$ evaluated questions,
\begin{equation}
\mathrm{EM} = \frac{100}{N} \sum_{i=1}^{N}
\mathbf{1}\{\mathcal{N}(h_i) = \mathcal{N}(r_i)\}.
\label{eq:em}
\end{equation}
Because any difference in wording makes an answer count as wrong, EM is much stricter than the classification accuracy or consensus-based VQA accuracy reported in some studies, and its values should not be compared with them.

\paragraph{Token-level F1}
Token F1 relaxes EM by giving partial credit for shared words, regardless of their order \citep{rajpurkar2016squad}. Normalized answers were split on whitespace and matched as multisets, so that repeated words are counted. Let $H_i$ and $R_i$ be the token sequences of the prediction and the reference, and $c_H(t)$ and $c_R(t)$ the frequencies of token $t$ in each. The overlap $o_i$, precision $P_i$, recall $R_i^{\mathrm{tok}}$, and F1 are
\begin{align}
o_i &= \sum_t\min\{c_H(t),c_R(t)\}, \quad
P_i = \frac{o_i}{|H_i|}, \\
R_i^{\mathrm{tok}} &= \frac{o_i}{|R_i|}, \quad
F_i^{\mathrm{tok}} = \frac{2P_iR_i^{\mathrm{tok}}}{P_i+R_i^{\mathrm{tok}}},
\end{align}
and we report $100\,N^{-1} \sum_i F^{\mathrm{tok}}_i$. An answer with no shared words scores zero; if both normalized answers are empty, the score is one, and if only one is empty, it is zero.

\paragraph{ROUGE-L F1}
ROUGE-L adds sensitivity to word order \citep{lin2004rouge}. It is based on the longest common subsequence of the two answers, that is, the longest sequence of words that appears in both in the same order, though not necessarily adjacently. We used \texttt{rouge-score} version 0.1.2, which, after the normalization described above, tokenizes on alphanumeric characters and applies Porter stemming to words longer than three characters. Writing $\tilde{H}_i$ and $\tilde{R}_i$ for the resulting sequences and $L_i$ for the length of their longest common subsequence, precision is $L_i/|\tilde{H}_i|$ and recall is $L_i/|\tilde{R}_i|$. We report their harmonic mean (ROUGE-L F1), averaged over all questions, rather than ROUGE-L recall or ROUGE-Lsum.

\paragraph{Corpus BLEU-4}
BLEU measures the overlap of consecutive word $n$-grams ($n = 1,\dots,4$) between the generated answers and their references, aggregated over the evaluation set, and penalizes outputs that are too short \citep{papineni2002bleu}. We computed it with SacreBLEU version 2.5.1 \citep{post2018call}, using one reference per prediction, the \texttt{13a} tokenizer, case-sensitive scoring, exponential smoothing, and effective order disabled; no answer pairs were excluded. For each order, candidate $n$-gram matches were clipped to the counts in the corresponding reference and summed over all answer pairs, with no $n$-grams crossing answer boundaries. The four resulting modified precisions $\tilde{p}_n$ (smoothed when a count is zero) receive equal weights:
\begin{equation}
\mathrm{BLEU\text{-}4} = 100 \cdot \mathrm{BP} \cdot
\exp\!\left(\frac{1}{4}\sum_{n=1}^{4}\log \tilde{p}_n\right),
\label{eq:bleu}
\end{equation}
where the brevity penalty BP depends on the total candidate length $c$ and
reference length $r$:
\begin{equation}
\mathrm{BP} =
\begin{cases}
1, & c \ge r,\\
\exp(1 - r/c), & 0 < c < r,\\
0, & c = 0.
\end{cases}
\label{eq:bp}
\end{equation}

\paragraph{BERTScore F1}
The metrics above all reward matching words. BERTScore instead measures semantic similarity, giving credit to answers that convey the same meaning in different words \citep{zhang2019bertscore}. We computed it with \texttt{bert-score} using a locally cached \texttt{roberta-large} encoder, with representations from layer 17, a batch size of 64, inputs truncated at 512 tokens, and both IDF weighting and baseline rescaling disabled. Let $\mathbf{u}_a$ and $\mathbf{v}_b$ be the normalized contextual embeddings of the candidate and reference subword tokens. Precision averages, over the candidate tokens, the highest cosine similarity to any reference token, and recall does the same in reverse:
\begin{equation}
P^{B}_i = \frac{1}{m_i} \sum_{a=1}^{m_i} \max_{b}\, \mathbf{u}_a^{\top}\mathbf{v}_b,
\quad
R^{B}_i = \frac{1}{n_i} \sum_{b=1}^{n_i} \max_{a}\, \mathbf{u}_a^{\top}\mathbf{v}_b,
\label{eq:bertscore_pr}
\end{equation}
where padding and boundary special tokens are excluded from the averages. We
report
\begin{equation}
\mathrm{BERTScore\text{-}F1} = \frac{100}{N} \sum_{i=1}^{N}
\frac{2 P^{B}_i R^{B}_i}{P^{B}_i + R^{B}_i}.
\label{eq:bertscore_f1}
\end{equation}
Because baseline rescaling is disabled, the values are raw similarities, which fall in a narrow, high range even for weakly related text. They should therefore be read as relative comparisons between configurations rather than as percentages of correct answers.

\paragraph{Zero-shot baseline and category-level analysis}
Two further analyses place these metrics in context. First, to measure the effect of domain adaptation, we evaluated the original Qwen3-VL-2B-Instruct and LLaVA-1.5-7B checkpoints on the same evaluation sets without agricultural fine-tuning or in-context examples. Each model received the image, the question, and the corresponding domain instruction. Second, to identify which types of questions are most affected by quantization, we grouped question-answer pairs by category: nine categories for PlantExpertVQA (Table~\ref{tab:plantexpert_categories}) and seven for IP-VQA
(Table~\ref{tab:ipvqa_categories}). Within each category, the per-question metrics were averaged over its questions, and BLEU-4 was recomputed from that category's answer pairs alone. Overall results were computed from all evaluated questions rather than as an unweighted mean of category scores, and questions sharing an image were treated as separate observations.

\subsubsection{Deployment efficiency}
\label{sec:efficiency}
For the plant disease task, we deployed all five configurations of each model, namely the 16-bit fine-tuned model and its four quantized variants, giving ten served models in total. For the insect pest task, we deployed two configurations as a cross-domain check; their weight footprint and latency closely matched those of the corresponding plant disease configurations, so the remaining insect pest variants were not deployed separately. All reported efficiency figures therefore come from on-device measurements of the plant disease models. All ten configurations were served with identical settings from the same container image, with a maximum model length of 2,048 tokens and a GPU memory utilization of 0.60.
 
We report the weight footprint, which characterizes memory, together with three latencies that follow the NVIDIA NIM benchmarking definitions~\citep{nvidia_nim_metrics}: time to first token (TTFT), end-to-end latency (E2E), and inter-token latency (ITL).

\paragraph{Weight footprint} We define the weight footprint as the device memory occupied by the loaded model parameters, taken from the log that vLLM reports when model loading completes. It depends only on the model and its quantization format. We deliberately exclude the memory that the engine allocates afterwards, such as the KV cache and the CUDA graph workspace, because these depend on the memory allocation settings and the required concurrency.

\paragraph{Latency}
 
\begin{figure}[pos=htbp]
  \centering
  \includegraphics[width=\columnwidth]{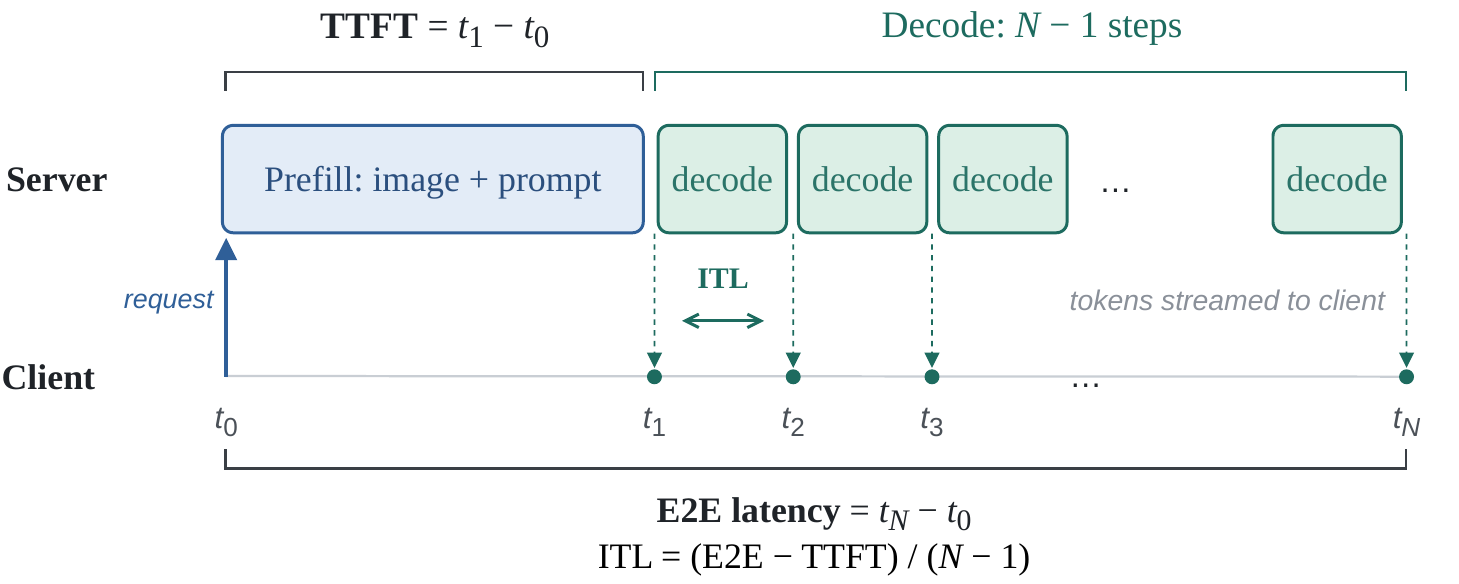}
  \caption{Illustration of the time to first token (TTFT) and inter-token latency (ITL) metrics, following \citep{nvidia_nim_metrics}.}
  \label{fig:ttft_itl}
\end{figure}

 
All latencies were measured one request at a time, with no other requests running. Each request paired one image with the question “What disease affects this plant?”, drawn from 40 randomly sampled unique images from the plant-disease test set. Every configuration received the same 40 images in the same order, and caching was disabled so that each request began a fresh cycle. Three warm-up requests were run before each test and excluded from the measurements. We capped every answer at exactly $N = 64$ tokens so that output length was identical across configurations and end-to-end latencies are directly comparable.
Writing $t_0$ for the time of submission, $t_1$ for the arrival of the first generated token, and $t_N$ for the end of the response,
\begin{equation}
\mathrm{TTFT} = t_1 - t_0, \qquad
\mathrm{E2E} = t_N - t_0, \qquad
\mathrm{ITL} = \frac{\mathrm{E2E} - \mathrm{TTFT}}{N - 1}.
\label{eq:latency}
\end{equation}
As illustrated in Figure~\ref{fig:ttft_itl}, TTFT is the delay before the user sees any output: it covers request handling and the prefill stage, in which the image is encoded and the prompt is processed. End-to-end latency (E2E) is the time to the final token, spanning the whole request. Inter-token latency (ITL), also called time per output token, is the average gap between consecutive tokens once generation is under way. The first token marks the end of prefill rather than a decoding step, so ITL averages over the remaining $N - 1$ tokens, isolating the cost of decoding from the one-time cost of prefill. The three metrics combine as $\mathrm{TTFT} + (n - 1)\,\mathrm{ITL}$, the time to produce an answer of $n$ tokens. We report all three as means over the 40 requests, with standard deviations for TTFT and E2E. The reciprocal of the mean ITL gives the per-user decoding rate in tokens per second, a restatement of ITL rather than a separate measurement.
 
\section{Results}
\label{sec:results}
We report results following the stages of the PlantVLM-Edge framework. We first examine how domain adaptation changes answer quality relative to the zero-shot backbones (Section~\ref{sec:res-adaptation}), then how post-training quantization affects answer quality on both datasets (Section~\ref{sec:res-quantization}), and then the memory and latency of each configuration measured on the Jetson AGX Orin (Section~\ref{sec:res-efficiency}). Next, we combine the two to analyze the accuracy-efficiency trade-off and identify recommended configurations (Section~\ref{sec:res-tradeoff}). Finally, we present an example interaction with the deployed system using the recommended configuration (Section~\ref{sec:res-demo}).

\subsection{Effect of domain adaptation}
\label{sec:res-adaptation}
Table~\ref{tab:answer_quality} reports answer quality for the zero-shot backbones, the fine-tuned 16-bit models, and their four quantized variants on both datasets. Domain adaptation produces by far the largest change in answer quality in this study. On PlantExpertVQA, fine-tuning raised ROUGE-L from 15.45 to 78.18 for Qwen3-VL-2B and from 16.11 to 78.19 for LLaVA-1.5-7B, and exact match from below 2\% to about 34\% for both models. On IP-VQA, it raised ROUGE-L from 30.45 to 54.67 for Qwen3-VL-2B and from 26.25 to 54.41 for LLaVA-1.5-7B. The zero-shot models therefore answered in a form that differed substantially from the reference answers, and adaptation was essential for both backbones.

After fine-tuning, the two backbones reached almost identical answer quality: 78.18 against 78.19 ROUGE-L on PlantExpertVQA and 54.67 against 54.41 on IP-VQA, despite LLaVA-1.5-7B having more than three times as many parameters. 
Absolute scores are lower on IP-VQA than on PlantExpertVQA throughout, at roughly 54 against 78 ROUGE-L, and exact match stays below 5\%. This reflects the less templated answer distribution of IP-VQA, in which longer free-form descriptions make exact string agreement rare. BERTScore F1 changed much less with fine-tuning than the other metrics (for example, from 85.03 to 96.56 for Qwen3-VL-2B on PlantExpertVQA). This is largely a property of unrescaled BERTScore, which occupies a narrow, high range even for weakly related text (Section~\ref{sec:answer-quality}); its values are therefore most informative as relative comparisons between configurations.

\begin{table*}[!t]
\centering
\small
\caption{Answer quality of both backbones on the PlantExpertVQA and the IP-VQA test set. All metrics are reported on a 0-100 scale; higher is better. For each model, the best result among the fine-tuned configurations in each column is in bold.}
\label{tab:answer_quality}
\setlength{\tabcolsep}{4pt}
\begin{tabular}{@{}l rrrrr c rrrrr@{}}
\toprule
 & \multicolumn{5}{c}{\textbf{PlantExpertVQA}} & & \multicolumn{5}{c}{\textbf{IP-VQA}} \\
\cmidrule(lr){2-6} \cmidrule(lr){8-12}
\textbf{Configuration} & \textbf{R-L} & \textbf{F1} & \textbf{EM} & \textbf{BERT} & \textbf{BLEU}
 & & \textbf{R-L} & \textbf{F1} & \textbf{EM} & \textbf{BERT} & \textbf{BLEU} \\
\midrule
\multicolumn{12}{@{}l}{\textit{Qwen3-VL-2B}} \\
Zero-shot          & 15.45 & 19.35 & 0.29 & 85.03 & 3.10 & & 30.45 & 33.22 & 0.33 & 89.35 & 5.70 \\
Fine-tuned (BF16)  & \textbf{78.18} & \textbf{79.52} & \textbf{33.64} & \textbf{96.56} & 65.87
                   & & 54.67 & 57.13 & \textbf{4.91} & \textbf{94.06} & 30.03 \\
W8A16-RTN          & 78.04 & 79.42 & 33.53 & 96.55 & 65.73
                   & & \textbf{54.70} & \textbf{57.18} & 4.87 & \textbf{94.06} & \textbf{30.05} \\
W8A8-SmoothQuant   & 78.11 & 79.46 & 33.60 & 96.55 & \textbf{65.90}
                   & & 54.40 & 56.84 & 4.57 & 94.02 & 29.58 \\
W4A16-RTN          & 76.44 & 77.82 & 30.79 & 96.25 & 64.04
                   & & 52.95 & 55.39 & 4.18 & 93.81 & 28.38 \\
W4A16-AWQ          & 77.11 & 78.50 & 31.95 & 96.36 & 64.28
                   & & 53.91 & 56.37 & 4.38 & 93.94 & 29.13 \\
\midrule
\multicolumn{12}{@{}l}{\textit{LLaVA-1.5-7B}} \\
Zero-shot          & 16.11 & 19.54 & 1.92 & 85.96 & 3.60 & & 26.25 & 29.73 & 0.31 & 89.57 & 5.06 \\
Fine-tuned (FP16)  & \textbf{78.19} & \textbf{79.50} & \textbf{33.73} & \textbf{96.57} & \textbf{66.00}
                   & & 54.41 & 56.86 & 4.76 & \textbf{94.03} & 29.81 \\
W8A16-RTN          & 78.17 & 79.49 & 33.72 & \textbf{96.57} & 65.99
                   & & \textbf{54.43} & \textbf{56.89} & \textbf{4.77} & \textbf{94.03} & \textbf{29.84} \\
W8A8-SmoothQuant   & 78.11 & 79.41 & 33.62 & 96.56 & 65.46
                   & & \textbf{54.43} & 56.78 & 4.57 & \textbf{94.03} & 29.19 \\
W4A16-RTN          & 77.99 & 79.26 & 33.58 & 96.53 & 65.79
                   & & 54.20 & 56.67 & 4.50 & 94.00 & 29.47 \\
W4A16-AWQ          & 77.86 & 79.22 & 33.36 & 96.51 & 65.74
                   & & 54.00 & 56.51 & 4.36 & 93.98 & 29.45 \\
\bottomrule
\end{tabular}

\vspace{2pt}
\begin{minipage}{\textwidth}
\footnotesize
R-L: ROUGE-L F1; F1: token-level F1; EM: normalized exact match; BERT: BERTScore F1; BLEU: corpus BLEU-4.
\end{minipage}
\end{table*}

\subsection{Answer quality under quantization}
\label{sec:res-quantization}
We next compare each quantized configuration with its fine-tuned 16-bit counterpart, which serves as the quality reference. Tables~\ref{tab:cat_plant} and~\ref{tab:cat_ip} report category-level ROUGE-L, giving the 16-bit score and the change under each quantized configuration, so that losses can be read directly.

\subsubsection{PlantExpertVQA}
\label{sec:res-quant-plant}

\paragraph{Qwen3-VL-2B}
Both 8-bit configurations closely matched the fine-tuned model on all five metrics (Table~\ref{tab:answer_quality}): W8A16-RTN and W8A8-SmoothQuant scored 78.04 and 78.11 ROUGE-L against 78.18, and SmoothQuant marginally exceeded the reference on BLEU-4. At 4 bits, a small cost appeared, and the quantization method began to matter. W4A16-AWQ retained 98.6\% of full-precision ROUGE-L, against 97.8\% for W4A16-RTN, and outperformed RTN on all five metrics. Going from full precision to W4A16-RTN, ROUGE-L fell by 1.74 points and BERTScore F1 by only 0.31, whereas exact match fell by 2.85 points, or 8.5\% of its value against 2.2\% for ROUGE-L. Exact match is the most sensitive metric because any change in wording makes the whole answer count as wrong, whereas ROUGE-L and BERTScore give credit for partial and semantic overlap.

The category-level results (Table~\ref{tab:cat_plant}) show where this cost falls. At 8 bits, every category stayed within 0.6 ROUGE-L of full precision. At 4 bits, the loss was uneven: under round-to-nearest, it ranged from 0.28 points in Visual Attribute Grounding to 3.67 points in Comprehensive Description, with Plant Species Identification next at 2.36. AWQ recovered most of this loss in Comprehensive Description (from 3.67 to 0.93 points) and Counterfactual Reasoning (from 1.84 to 0.41). However, it trailed round-to-nearest in four of the nine categories, by between 0.03 and 1.58 points, most notably in Visual Attribute Grounding (a loss of 1.86 points against 0.28) and Existence \& Sanity Check (1.57 against 0.32). The benefit of AWQ is therefore concentrated in long, explanatory answers, which matter most for a conversational diagnostic system.

\paragraph{LLaVA-1.5-7B}
Quantization was close to lossless across the entire range for the larger backbone (Table~\ref{tab:answer_quality}). Every configuration stayed within 0.33 ROUGE-L of full precision, BERTScore F1 varied by at most 0.06 points, and exact match by at most 0.37. W8A16-RTN was effectively identical to the fine-tuned model, and at 4 bits round-to-nearest scored slightly higher than AWQ (77.99 against 77.86). The category-level results (Table~\ref{tab:cat_plant}) were equally flat: no configuration fell more than 0.73 ROUGE-L below full precision in any category, the largest deviation being a 0.73-point loss in Counterfactual Reasoning under W4A16-AWQ.

Although the two backbones reached the same quality at full precision, they differed in their sensitivity to 4-bit quantization: Qwen3-VL-2B lost 1.07 (AWQ) and 1.74 (RTN) ROUGE-L points, whereas LLaVA-1.5-7B lost only 0.33 and 0.20.

\subsubsection{IP-VQA}
\label{sec:res-quant-ip}

\paragraph{Qwen3-VL-2B}
The 8-bit configurations again preserved answer quality (Table~\ref{tab:answer_quality}). W8A16-RTN scored 54.70 ROUGE-L against 54.67 at full precision, a negligible difference with identical BERTScore F1 (94.06), and W8A8-SmoothQuant scored 54.40. At 4 bits, the two methods separated more clearly than on PlantExpertVQA. Round-to-nearest lost 1.72 ROUGE-L, similar to its loss on the plant task in absolute terms, whereas AWQ lost only 0.76, retaining 98.6\% of full-precision ROUGE-L, and outperformed round-to-nearest on all five metrics. AWQ's lead over round-to-nearest was therefore wider than on PlantExpertVQA (0.96 against 0.67 points), and it recovered 56\% of the round-to-nearest loss on IP-VQA compared with 39\% on PlantExpertVQA.

This difference is consistent with the structure of the two datasets. PlantExpertVQA contains many repeated question and answer patterns, whereas IP-VQA has more varied questions and responses. Repetitive, templated answers may be less sensitive to coarse weight rounding, whereas more varied answers make the effect of quantization more visible, which also leaves more room for AWQ to recover. The category-level results (Table~\ref{tab:cat_ip}) show that the 4-bit degradation was broad rather than concentrated: under round-to-nearest, every category lost between 0.9 and 2.4 points, with Pest Control Methods the most affected ($-2.43$). AWQ recovered part of this loss in six of the seven categories. The exception was Optimal Timing, the smallest category ($n = 345$), where AWQ lost 1.87 points against 0.98 for round-to-nearest.

\paragraph{LLaVA-1.5-7B}
Quantization left answer quality essentially unchanged for the larger backbone (Table~\ref{tab:answer_quality}). Across all five configurations, ROUGE-L varied by 0.43 points and BERTScore F1 by 0.05. Both 8-bit configurations scored 54.43 ROUGE-L against 54.41 at full precision, with BERTScore F1 identical at 94.03. At 4 bits, round-to-nearest again scored slightly higher than AWQ (54.20 against 54.00). The category-level results (Table~\ref{tab:cat_ip}) were equally flat. Under 4-bit round-to-nearest, no category fell more than 0.9 ROUGE-L below full precision, compared with losses of 0.9 to 2.4 points for Qwen3-VL-2B on the same dataset. AWQ trailed round-to-nearest in six of the seven categories, most in Pest Identification ($-0.69$ against $-0.31$), and SmoothQuant recorded the highest score in four of the seven categories by small margins.

\subsubsection{Summary across models and domains}
\label{sec:res-quant-summary}
The four model-dataset combinations give a consistent picture. Eight-bit quantization changed aggregate ROUGE-L by at most 0.3 points in every case, regardless of method. Four-bit quantization cost between 0.3\% and 3.1\% of ROUGE-L, with the larger losses falling on the compact backbone and on the less templated IP-VQA dataset. The choice of 4-bit method mattered only where this loss was material: for Qwen3-VL-2B, AWQ outperformed round-to-nearest in both domains and retained 98.6\% of full-precision ROUGE-L on each, whereas for LLaVA-1.5-7B the two methods differed by at most 0.2 points, with round-to-nearest slightly ahead.

\begin{table*}[!t]
\centering
\small
\caption{Category-level ROUGE-L on the PlantExpertVQA test set. For each backbone, the first column gives the 16-bit score and the remaining columns give the change under each quantized configuration.}
\label{tab:cat_plant}
\setlength{\tabcolsep}{3.5pt}
\begin{tabular}{@{}l r rrrrr c rrrrr@{}}
\toprule
 & & \multicolumn{5}{c}{\textbf{Qwen3-VL-2B}} & & \multicolumn{5}{c}{\textbf{LLaVA-1.5-7B}} \\
\cmidrule(lr){3-7} \cmidrule(lr){9-13}
\textbf{Category} & $\mathbf{n}$ & \textbf{16-bit} & \textbf{W8} & \textbf{SQ} & \textbf{W4} & \textbf{AWQ}
 & & \textbf{16-bit} & \textbf{W8} & \textbf{SQ} & \textbf{W4} & \textbf{AWQ} \\
\midrule
Causal Reasoning & 2,180 & 66.02 & $-0.26$ & $-0.58$ & $\mathbf{-1.41}$ & $\mathbf{-1.44}$ & & 65.74 & $-0.03$ & $-0.09$ & $-0.52$ & $-0.25$ \\
Comprehensive Descr. & 1,767 & 64.74 & $-0.21$ & $+0.41$ & $\mathbf{-3.67}$ & $-0.93$ & & 63.75 & $+0.05$ & 0.00 & $-0.19$ & $-0.26$ \\
Counterfactual Reas. & 1,202 & 74.00 & $-0.15$ & $+0.26$ & $\mathbf{-1.84}$ & $-0.41$ & & 74.58 & $-0.06$ & $-0.11$ & $+0.41$ & $-0.73$ \\
Detailed Verification & 1,689 & 79.79 & $-0.14$ & $+0.08$ & $-0.65$ & $-0.79$ & & 80.38 & 0.00 & $-0.20$ & $-0.24$ & $-0.51$ \\
Existence \& Sanity & 72 & 100.00 & 0.00 & 0.00 & $-0.32$ & $\mathbf{-1.57}$ & & 100.00 & 0.00 & 0.00 & 0.00 & 0.00 \\
General Health & 1,337 & 97.18 & $-0.03$ & $-0.07$ & $-0.69$ & $-0.30$ & & 97.28 & $+0.01$ & $-0.19$ & $-0.33$ & $-0.14$ \\
Plant Species ID & 2,147 & 89.38 & $-0.05$ & $-0.23$ & $\mathbf{-2.36}$ & $\mathbf{-1.80}$ & & 89.48 & $-0.06$ & $-0.09$ & $-0.38$ & $-0.26$ \\
Specific Disease ID & 3,972 & 76.87 & $-0.12$ & $-0.02$ & $\mathbf{-1.80}$ & $-0.95$ & & 77.09 & $-0.02$ & $-0.04$ & 0.00 & $-0.36$ \\
Visual Attribute Gr. & 716 & 87.56 & $-0.13$ & $-0.16$ & $-0.28$ & $\mathbf{-1.86}$ & & 86.98 & $+0.01$ & $+0.19$ & $-0.58$ & $-0.07$ \\
\bottomrule
\end{tabular}

\vspace{2pt}
\begin{minipage}{\textwidth}
\footnotesize
16-bit: ROUGE-L of the fine-tuned model (BF16 for Qwen3-VL-2B, FP16 for LLaVA-1.5-7B). W8: W8A16-RTN; SQ: W8A8-SmoothQuant; W4: W4A16-RTN; AWQ: W4A16-AWQ. Quantized columns give the change in ROUGE-L points relative to 16-bit; losses of 1.0 point or more are in bold.
\end{minipage}
\end{table*}

\begin{table*}[!t]
\centering
\small
\caption{Category-level ROUGE-L on the IP-VQA test set. For each backbone, the first column gives the 16-bit score and the remaining columns give the change under each quantized configuration.}
\label{tab:cat_ip}
\setlength{\tabcolsep}{3.5pt}
\begin{tabular}{@{}l r rrrrr c rrrrr@{}}
\toprule
 & & \multicolumn{5}{c}{\textbf{Qwen3-VL-2B}} & & \multicolumn{5}{c}{\textbf{LLaVA-1.5-7B}} \\
\cmidrule(lr){3-7} \cmidrule(lr){9-13}
\textbf{Category} & $\mathbf{n}$ & \textbf{16-bit} & \textbf{W8} & \textbf{SQ} & \textbf{W4} & \textbf{AWQ}
 & & \textbf{16-bit} & \textbf{W8} & \textbf{SQ} & \textbf{W4} & \textbf{AWQ} \\
\midrule
Optimal Timing & 345 & 53.40 & $+0.17$ & $+0.29$ & $-0.98$ & $\mathbf{-1.87}$ & & 52.64 & $-0.22$ & $+0.79$ & $-0.85$ & $-0.50$ \\
Other & 3,562 & 58.34 & $+0.11$ & $-0.12$ & $\mathbf{-1.97}$ & $-0.62$ & & 57.94 & $-0.03$ & $+0.27$ & $-0.17$ & $-0.25$ \\
Pest Control Methods & 3,234 & 47.40 & $+0.12$ & $-0.15$ & $\mathbf{-2.43}$ & $-0.43$ & & 47.27 & $+0.10$ & $+0.07$ & $-0.01$ & $-0.18$ \\
Pest Feeding \& Damage & 4,135 & 48.55 & $-0.03$ & $-0.38$ & $\mathbf{-1.65}$ & $-0.97$ & & 48.57 & $-0.03$ & $-0.01$ & $-0.29$ & $-0.43$ \\
Pest Identification & 4,354 & 62.77 & $+0.07$ & $-0.32$ & $\mathbf{-1.34}$ & $-0.80$ & & 62.54 & $+0.08$ & $-0.31$ & $-0.31$ & $-0.69$ \\
Pest Monitoring & 1,121 & 55.80 & $+0.06$ & $-0.04$ & $\mathbf{-1.23}$ & $-0.99$ & & 55.36 & $+0.05$ & $+0.09$ & $-0.22$ & $-0.48$ \\
Visual / Characteristics & 2,199 & 54.47 & $-0.21$ & $-0.51$ & $\mathbf{-1.46}$ & $-0.65$ & & 53.84 & $+0.06$ & $+0.11$ & $-0.06$ & $-0.33$ \\
\bottomrule
\end{tabular}

\vspace{2pt}
\begin{minipage}{\textwidth}
\footnotesize
16-bit: ROUGE-L of the fine-tuned model (BF16 for Qwen3-VL-2B, FP16 for LLaVA-1.5-7B). W8: W8A16-RTN; SQ: W8A8-SmoothQuant; W4: W4A16-RTN; AWQ: W4A16-AWQ. Quantized columns give the change in ROUGE-L points relative to 16-bit; losses of 1.0 point or more are in bold.
\end{minipage}
\end{table*}

\subsection{On-device efficiency}
\label{sec:res-efficiency}
Table~\ref{tab:efficiency} reports the weight footprint and the three latency metrics measured on the Jetson AGX Orin for the ten plant disease configurations.

\begin{table}[!t]
\centering
\footnotesize
\caption{On-device efficiency on the Jetson AGX Orin for the plant disease models. Latencies are means over 40 requests of 64 tokens each. Standard deviations of TTFT and E2E are in parentheses. BF16 and FP16 denote the fine-tuned 16-bit models; SQ: SmoothQuant. The lowest value for each model is in bold.}
\label{tab:efficiency}
\setlength{\tabcolsep}{2.5pt}
\begin{tabular}{@{}l r r r r@{}}
\toprule
\textbf{Config.} & \textbf{Weights} & \textbf{TTFT} & \textbf{E2E} & \textbf{ITL} \\
 & \textbf{(GiB)} & \textbf{(ms)} & \textbf{(ms)} & \textbf{(ms/tok)} \\
\midrule
\multicolumn{5}{@{}l}{\textit{Qwen3-VL-2B}} \\
BF16               & 3.96 & 130.4\,(30) & 1535.6\,(31) & 22.3 \\
W8A16-RTN          & 2.65 & 123.3\,(31) & 1042.2\,(32) & 14.6 \\
W8A8-SQ            & 2.65 & \textbf{116.1}\,(29) & 1117.5\,(30) & 15.9 \\
W4A16-RTN          & \textbf{2.01} & 124.1\,(32) & 782.5\,(33) & 10.5 \\
W4A16-AWQ          & \textbf{2.01} & 123.4\,(31) & \textbf{777.7}\,(32) & \textbf{10.4} \\
\midrule
\multicolumn{5}{@{}l}{\textit{LLaVA-1.5-7B}} \\
FP16               & 13.16 & 463.8\,(14) & 5451.8\,(14) & 79.2 \\
W8A16-RTN          & 7.13  & 419.2\,(14) & 3235.9\,(14) & 44.7 \\
W8A8-SQ            & 7.13  & \textbf{248.0}\,(14) & 3136.7\,(14) & 45.9 \\
W4A16-RTN          & \textbf{4.20} & 374.8\,(14) & 1983.5\,(14) & \textbf{25.5} \\
W4A16-AWQ          & \textbf{4.20} & 359.0\,(7)  & \textbf{1968.5}\,(7) & \textbf{25.5} \\
\bottomrule
\end{tabular}
\end{table}

\subsubsection{Memory and decoding speed}
\label{sec:res-memory}
Bit-width determined both the weight footprint and the decoding speed (Table~\ref{tab:efficiency}, Weights and ITL columns). Configurations at the same bit-width occupied identical memory, and the two 4-bit configurations decoded within 0.1~ms per token of each other. The exception was W8A8-SmoothQuant, which decoded 1.2-1.3~ms per token more slowly than W8A16-RTN, consistent with the added cost of quantizing activations at every decoding step. Relative to 16 bits, the weight footprint fell by 33\% (8 bits) and 49\% (4 bits) for Qwen3-VL-2B, and by 46\% and 68\% for LLaVA-1.5-7B. The compact backbone compresses less because its unquantized components, namely the vision encoder, embeddings, and output head, make up a larger share of its parameters.

Decoding accelerated accordingly, with inter-token latency falling by 2.14$\times$ for Qwen3-VL-2B and 3.11$\times$ for LLaVA-1.5-7B. Each generated token requires the language-backbone weights to be read from memory, so fewer weight bits mean less memory traffic and faster token generation. The decoding speed-up does not exactly track the footprint reduction, because the footprint includes components that play different roles during decoding. The 16-bit vision encoder occupies memory but is not used after prefill, which is why Qwen3-VL-2B decoded 2.14$\times$ faster even though its footprint shrank by only 1.97$\times$. The 16-bit output head, in contrast, is read at every decoding step and limits the attainable speed-up.

\subsubsection{Time to first token}
\label{sec:res-ttft}
Before a model can emit its first token, it must encode the image and process the prompt, a stage known as prefill (Figure~\ref{fig:ttft_itl}). Quantization helps less here for two reasons: the vision encoder, which runs entirely during prefill, remains at 16 bits in every configuration, and prefill processes all input tokens in parallel, so it is limited by computation rather than by the memory traffic that dominates decoding. For Qwen3-VL-2B, the four quantized configurations reached their first token in 116-124~ms, against 130~ms at 16 bits (Table~\ref{tab:efficiency}), differences smaller than the standard deviation across requests. LLaVA-1.5-7B started much more slowly, at 464~ms, because it processes 576 visual tokens against 128-256 for Qwen3-VL-2B and prefills them through a larger language backbone. Weight-only quantization reduced its TTFT by 10-23\%, whereas W8A8-SmoothQuant reduced it by 46\%, to 248~ms. This is the only case in which quantizing activations clearly helped, and it occurred in the model whose prefill is long enough for 8-bit arithmetic to pay off.

\subsubsection{End-to-end latency}
\label{sec:res-e2e}
Because every request generated exactly 64 tokens, end-to-end latencies in Table~\ref{tab:efficiency} are directly comparable across configurations. Quantization reduced the response time from 1.54 to 0.78~s for Qwen3-VL-2B (1.97$\times$) and from 5.45 to 1.97~s for LLaVA-1.5-7B (2.77$\times$), with W4A16-AWQ the fastest configuration for both models. Decoding accounted for more than 80\% of the response time in every configuration, so decoding speed, not TTFT, determined the ranking (Figure~\ref{fig:ttft_itl} illustrates the two phases). For example, W8A8-SmoothQuant gave LLaVA-1.5-7B its fastest first token (248 against 359~ms for W4A16-AWQ), but its full response took 3.14 against 1.97~s because it decoded 1.8$\times$ more slowly. As decoding becomes faster, TTFT accounts for a larger share of the response: for both models, it rises from 8.5\% at 16 bits to 15.9\% (Qwen3-VL-2B) and 18.2\% (LLaVA-1.5-7B) with W4A16-AWQ.

\subsubsection{Backbone comparison}
\label{sec:res-backbone}
The choice of backbone had a larger effect on deployment cost than the choice of quantization (Table~\ref{tab:efficiency}). Even unquantized, Qwen3-VL-2B occupied less weight memory than the most compressed LLaVA-1.5-7B configuration (3.96 against 4.20~GiB) and was faster than every LLaVA-1.5-7B configuration: it reached its first token in 130~ms against at best 248~ms, decoded at 22.3 against at best 25.5~ms per token, and completed a 64-token answer in 1.54 against at best 1.97~s. At equal precision the gap widened. With both models at W4A16-AWQ, Qwen3-VL-2B used 2.09$\times$ less memory, reached its first token 2.9$\times$ sooner, decoded 2.45$\times$ faster, and completed a 64-token answer 2.5$\times$ sooner. Quantization reduced inter-token latency by at most 3.1$\times$ within a model, which was not enough for any LLaVA-1.5-7B configuration to match the unquantized Qwen3-VL-2B. Because the two fine-tuned backbones reach the same answer quality (Section~\ref{sec:res-adaptation}), backbone selection was the dominant efficiency decision in this study and quantization the secondary one.

\begin{figure*}[pos=t]
    \centering
    \begin{minipage}[t]{0.49\textwidth}
        \centering
        \includegraphics[width=\linewidth]{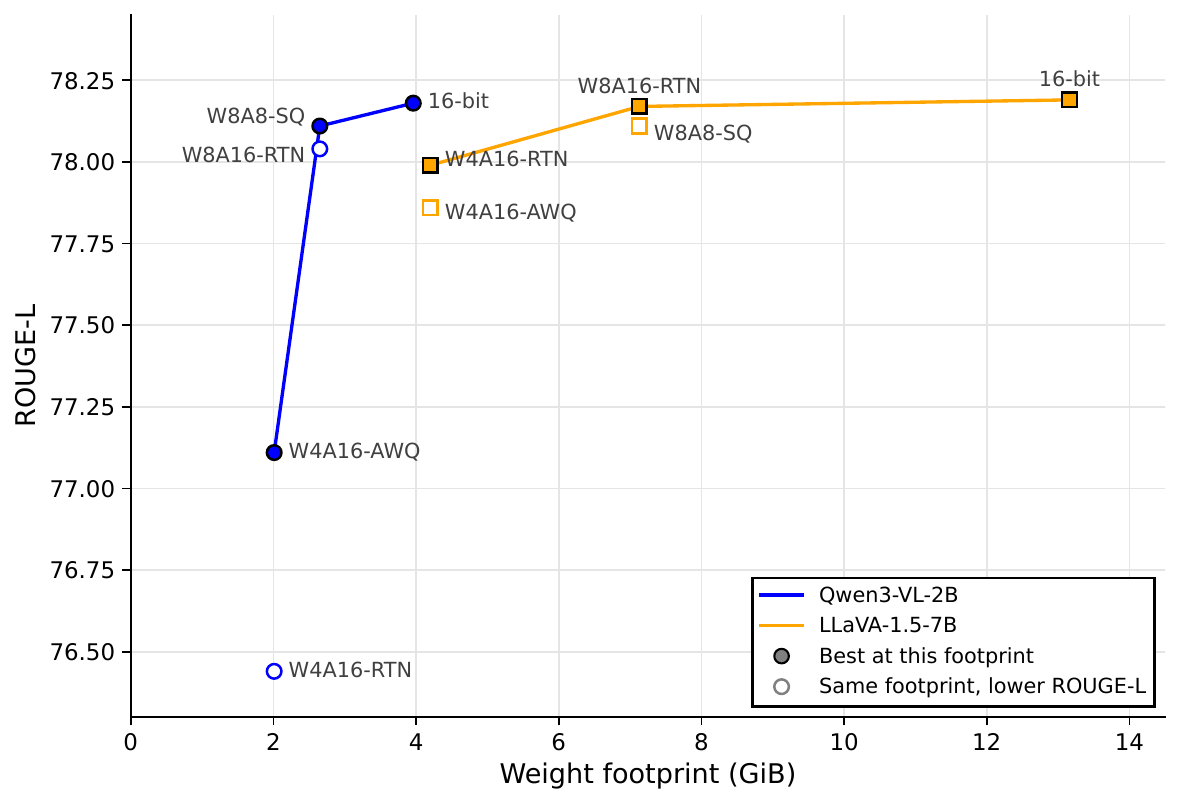}\\[2pt]
        {\small (a) Weight footprint}
    \end{minipage}\hfill
    \begin{minipage}[t]{0.49\textwidth}
        \centering
        \includegraphics[width=\linewidth]{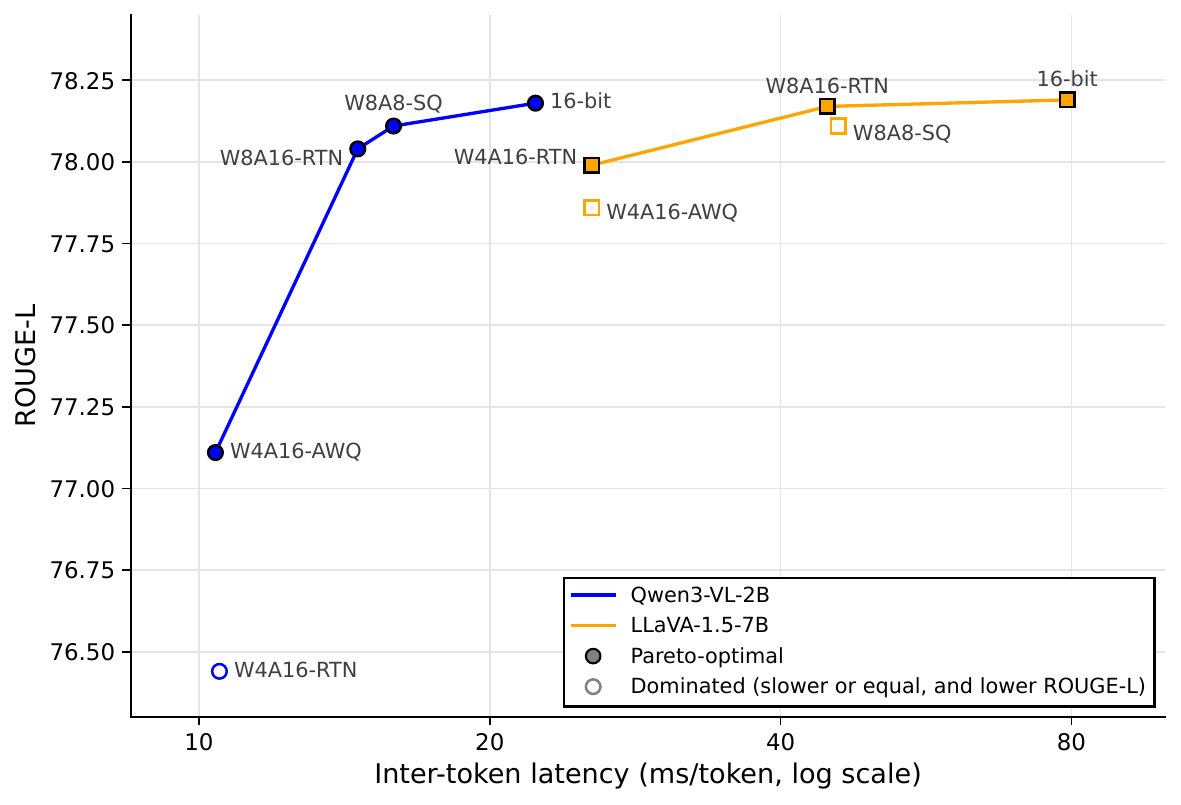}\\[2pt]
        {\small (b) Inter-token latency}
    \end{minipage}
    \caption{Accuracy-efficiency trade-off for the ten PlantExpertVQA configurations on the Jetson AGX Orin: ROUGE-L against (a) weight footprint and (b) inter-token latency. Lines connect the Pareto-optimal configurations of each model.}
    \label{fig:tradeoff}
\end{figure*}

\subsection{Accuracy-efficiency trade-off}
\label{sec:res-tradeoff}
Figure~\ref{fig:tradeoff} plots all ten configurations against weight footprint (a) and inter-token latency (b). Answer quality spans only 1.8 ROUGE-L points across the set, whereas memory and decoding speed differ by more than a factor of six. Efficiency, rather than answer quality, therefore separates the configurations.

\paragraph{Choice of backbone}
Because the two backbones reach equivalent answer quality, the comparison between them rests on efficiency, where Qwen3-VL-2B is the clear choice. At 8 bits, it remains within 0.2 ROUGE-L of every LLaVA-1.5-7B configuration while requiring about 37\% less memory, reaching its first token at least 2.0$\times$ sooner, and decoding at least 1.6$\times$ faster. Quantizing the larger model further does not change this conclusion, because even its 4-bit configurations remain more expensive than the compact model at 8 bits.

\paragraph{Choice of configuration}
Among the Qwen3-VL-2B configurations, three are dominated. The 16-bit model is outperformed by W8A16-RTN, which preserves its answer quality (99.8\% of full-precision ROUGE-L) while using 33\% less memory and decoding 1.53$\times$ faster, so full precision offers no practical advantage. W4A16-RTN is outperformed by W4A16-AWQ, which has an identical footprint and decoding speed but scores 0.67 ROUGE-L higher. W8A8-SmoothQuant lies marginally above W8A16-RTN at equal footprint (Figure~\ref{fig:tradeoff}a), but the 0.07-point difference is too small to be meaningful, and W8A16-RTN decodes 1.3~ms per token faster, completes a 64-token answer 75~ms sooner (Figure~\ref{fig:tradeoff}b), and requires no calibration data. SmoothQuant's one advantage is the lowest TTFT of the five Qwen3-VL-2B configurations, but over a 64-token answer it finishes 75~ms behind W8A16-RTN and 340~ms behind W4A16-AWQ.

\textit{Recommended configurations.}
Two configurations therefore remain. W8A16-RTN is essentially lossless, retaining 99.8\% of full-precision ROUGE-L on PlantExpertVQA and matching full precision on IP-VQA, and requires no calibration data. W4A16-AWQ gives up a further 1.2\% of ROUGE-L in exchange for a 24\% smaller footprint and 29\% lower inter-token latency than W8A16-RTN, requiring only 2.01~GiB of weights and completing a 64-token answer in 0.78~s. We therefore recommend Qwen3-VL-2B with W8A16-RTN as the default PlantVLM-Edge configuration, and W4A16-AWQ where memory is the binding constraint or where longer answers make decoding speed the priority.

\subsection{System demonstration}
\label{sec:res-demo}
\begin{figure*}[pos=t]
    \centering
    \includegraphics[height=0.55\textheight,keepaspectratio]{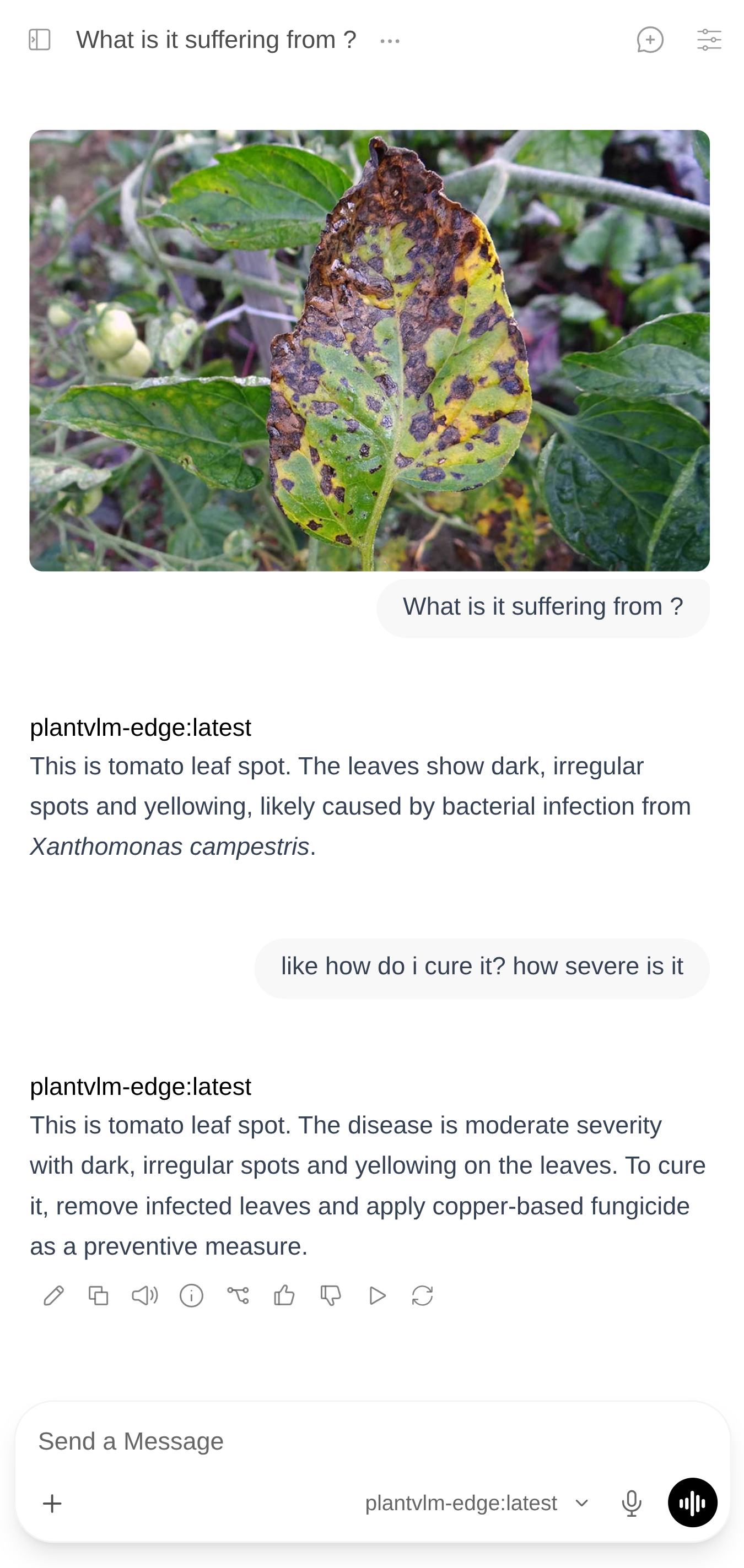}%
    \hspace{0.08\textwidth}%
    \includegraphics[height=0.55\textheight,keepaspectratio]{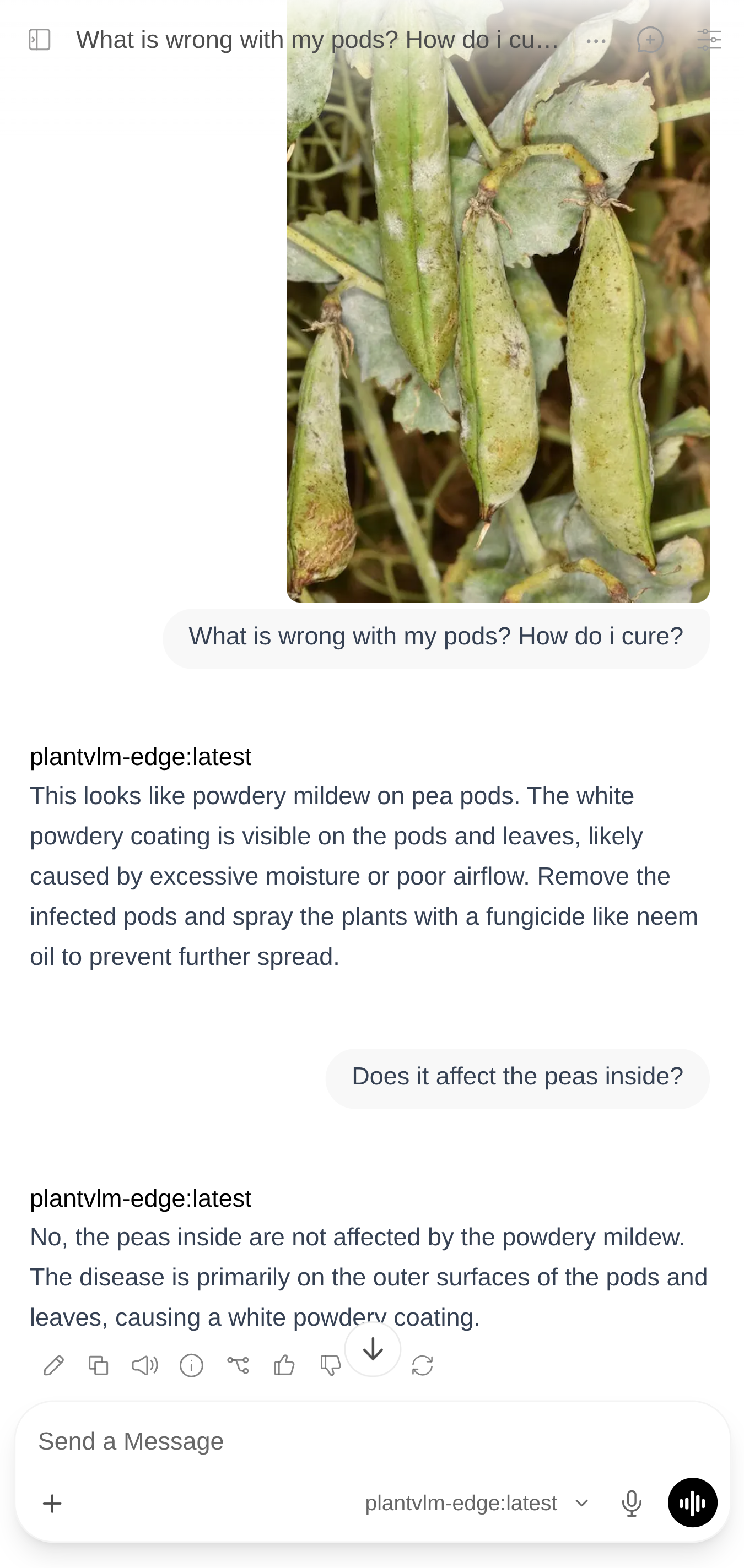}
    \caption{Screenshots of multi-turn conversations with the deployed PlantVLM-Edge system (Qwen3-VL-2B, W8A16-RTN) on the Jetson AGX Orin, accessed offline through a browser.}
    \label{fig:chat-demo}
\end{figure*}

Figure~\ref{fig:chat-demo} illustrates PlantVLM-Edge in use with the recommended Qwen3-VL-2B W8A16-RTN configuration. After a user uploads a photo of an infested leaf, the model identifies the problem and describes the visible damage, and answers a follow-up question about management in the same conversation, drawing on the earlier context. The interface runs entirely on the Jetson AGX Orin and is reached from a laptop or smartphone over the local network, with no internet connection. The image is taken from the test set.
Open WebUI also supports voice input, which could help users in the field whose hands are occupied. In the current deployment, however, speech recognition is not performed on the device, so the fully offline operation described here applies to text-based interaction; running speech recognition locally is left for future work.

\section{Discussion}
\label{sec:discussion}
The results show that a compact VLM, once adapted to the target domain, can match a model more than three times its size on agricultural question answering, and that post-training quantization then lowers its deployment cost further at little or no loss in answer quality. This section discusses what these findings imply for building edge-deployable diagnostic systems, translates them into deployment recommendations, and outlines the limitations of the present study.

\subsection{Domain adaptation and backbone selection}
\label{sec:disc-backbone}
Domain adaptation was the decisive step for answer quality. Fine-tuning raised ROUGE-L by 62.1-62.7 points on PlantExpertVQA and by 24.2-28.2 points on IP-VQA (Table~\ref{tab:answer_quality}), whereas no quantized configuration changed aggregate ROUGE-L by more than 1.8 points. After adaptation, the two backbones were indistinguishable in aggregate answer quality on both datasets, despite LLaVA-1.5-7B having more than three times as many parameters. Our fine-tuned Qwen3-VL-2B reached 78.18 ROUGE-L on PlantExpertVQA, compared with 61.70 reported by the dataset's authors for the same model fine-tuned on 500 images~\citep{sakib2026visual}. The figures are not directly comparable, because the evaluation subsets and training settings differ, but the gap is consistent with the larger training subset used here and supports the authors' observation that the dataset enables efficient domain adaptation.

For edge deployment, this reorders the usual priorities. Because the adapted backbones reach the same answer quality, the choice between them rests on efficiency, and there the compact model was smaller and faster at every bit-width, decoding 2.4-3.6$\times$ faster than LLaVA-1.5-7B at equal precision (Table~\ref{tab:efficiency}; Figure~\ref{fig:tradeoff}). The larger model's better tolerance of 4-bit quantization did not close this gap: even unquantized, Qwen3-VL-2B was smaller and faster than every LLaVA-1.5-7B configuration. Backbone selection was therefore the dominant efficiency decision in this study, and quantization the secondary one. Because the two backbones also differ in architecture and pretraining, we interpret this as evidence that a compact model is sufficient for these tasks, rather than as a general effect of model size. A controlled comparison, such as models of different sizes from the same family, would be needed to separate the effect of model size from that of architecture and pretraining.

\subsection{Bit-width versus quantization method}
\label{sec:disc-quantization}
Bit-width and quantization method played distinct roles. Bit-width determined deployment cost: configurations at the same precision had identical weight footprints (Table~\ref{tab:quant-configs}), and the two 4-bit configurations decoded at the same speed (Table~\ref{tab:efficiency}). The quantization method mattered mainly for answer quality, and only at 4 bits on the compact backbone, where AWQ recovered 39\% (PlantExpertVQA) to 56\% (IP-VQA) of the loss incurred by round-to-nearest. At 8 bits, every method was effectively lossless, and the data-free W8A16-RTN matched the calibrated SmoothQuant in answer quality. Activation quantization gave no decoding benefit under single-user serving and helped only the compute-bound prefill of the larger model.

The comparison between datasets carries a methodological implication. The larger relative loss under 4-bit round-to-nearest on IP-VQA, and the wider gap between AWQ and round-to-nearest there (Tables~\ref{tab:cat_plant} and~\ref{tab:cat_ip}), suggest that quantization studies on highly templated benchmarks may understate the cost of low-bit deployment on more varied question distributions. Since questions from growers in the field are likely to be more varied than either benchmark, evaluations of compressed agricultural models should include datasets with diverse, free-form answers.

\subsection{The role of the serving stack}
\label{sec:disc-serving}
The latency benefits of quantization depend on the software that executes the quantized model, not only on the bit-width. In our vLLM deployment, 4-bit weights reduced inter-token latency by 2.14$\times$ for Qwen3-VL-2B and by 3.11$\times$ for LLaVA-1.5-7B (Table~\ref{tab:efficiency}). In contrast, \citet{shin2026rethinking} found on Jetson Orin devices that 4-bit quantization of small VLMs could slow token generation because of the overhead of dequantizing weights. The same bit-width can therefore speed up or slow down generation depending on the quantization format and inference kernels, which is why PlantVLM-Edge evaluates complete configurations end to end on the target device rather than inferring speed from model size.

The results also identify where the next latency gains lie. Once decoding is faster, time to first token becomes a larger share of the response (Figure~\ref{fig:ttft_itl}): for a 64-token answer from Qwen3-VL-2B, it rises from 8.5\% at 16 bits to 15.9\% at 4 bits. Time to first token barely changed under quantization, because the vision encoder, which runs during prefill, remained at 16 bits in every configuration. Compressing the vision encoder is therefore the most direct next step for reducing latency, together with serving engines optimized for the Jetson platform, such as TensorRT-LLM.

\section{Conclusion}
\label{sec:conclusion}
This paper presented PlantVLM-Edge, a quantized vision-language framework for conversational crop health diagnosis that runs entirely on edge devices. The framework adapts a pretrained VLM to agricultural question answering with LoRA, compresses it with post-training quantization, and serves it on an NVIDIA Jetson AGX Orin through an offline, browser-based chat interface. We instantiated it with two backbones, Qwen3-VL-2B and LLaVA-1.5-7B, on plant disease (PlantExpertVQA) and insect pest (IP-VQA) diagnosis, and evaluated four quantization configurations for both answer quality and on-device efficiency.
Three main findings emerge. First, domain adaptation determines answer quality, whereas quantization costs little: fine-tuning raised ROUGE-L by 24-63 points, 8-bit quantization changed it by at most 0.3 points, and 4-bit AWQ retained at least 98.6\% of full-precision ROUGE-L for both backbones on both datasets. Second, bit-width determines memory and speed, whereas the quantization method matters only for answer quality at 4 bits on the compact backbone, where AWQ outperformed round-to-nearest. Third, backbone selection matters more than quantization: the fine-tuned Qwen3-VL-2B matched LLaVA-1.5-7B in answer quality while being smaller and faster at every bit-width. In its most compact form, PlantVLM-Edge requires 2.01~GiB of weights and completes a 64-token diagnostic answer in 0.78~s. We recommend Qwen3-VL-2B with W8A16-RTN as the default configuration and W4A16-AWQ where memory is limited. Future work will compress the vision encoder, evaluate Jetson-optimized serving engines and lower-cost devices such as the Jetson Orin Nano under multi-user load, and assess diagnostic correctness through expert evaluation and field trials with growers.

\section{Authorship Contribution}
\textbf{Gaurab Wagle}: Formal analysis, Data curation, Software, Writing - original draft; 
\textbf{Bishal Adhikari}: Formal analysis, Software, Writing - original draft; 
\textbf{Jiajia Li}: Conceptualization, Supervision, Writing - review \& editing;
\textbf{Xiaolei Huang}: Conceptualization, Writing - review \& editing;
\textbf{Dong Chen}: Conceptualization, Investigation, Supervision, Writing - review \& editing;

\section{ Declaration of AI-assisted Technologies}
\label{sec:AI-declaration}
During the preparation of this work, the author(s) used Generative AI (Claude) to assist with language editing, clarity, manuscript organization, and polishing the visual presentation of flowchart figures. After using this tool, the author(s) reviewed and edited the content and figures as needed and take(s) full responsibility for the content of the published article.


\bibliographystyle{cas-model2-names}
\bibliography{references}

\end{document}